\documentclass[10pt, a4paper]{article} 

\usepackage{authblk} 
\usepackage[T1]{fontenc} 
\usepackage[utf8]{inputenc} 
\usepackage{xfrac} 
\usepackage{subfiles} 
\usepackage{amsmath} %
\usepackage{makecell}
\usepackage{enumitem}
\usepackage{caption}
\usepackage{etoolbox}
\usepackage{layouts}
\usepackage{tabulary} 
\usepackage{pbox}
\usepackage{afterpage} 
\usepackage{lscape}
\usepackage[hyphens]{url}
\usepackage{hyperref}
\usepackage[noabbrev]{cleveref}
\usepackage{amssymb}
\usepackage{subfig}
\usepackage{algpseudocode}
\usepackage{algorithm}
\usepackage{multirow}

\hypersetup{breaklinks=true}

\setlist[description]{style=unboxed} 

\ifdef{\IEEEkeywords}{
	\let\keywords\IEEEkeywords	
}{}

\ifundef{\keywords}{
	\providecommand{\keywords}{\textbf{Keywords: }}	
}{}

\newcommand{\figurewidth}{\textwidth} 
\graphicspath{ {./Figures/} }

\newcolumntype{M}[1]{>{\centering\arraybackslash}{m}{#1}}

\begin{document}
	
\title{Visualizing and Exploring Dynamic High-Dimensional Datasets with LION-tSNE}
\author{Andrey Boytsov}
\author{Francois Fouquet}
\author{Thomas Hartmann}
\author{Yves LeTraon}
\affil{Interdisciplinary Centre for Security, Reliability and Trust\\
University of Luxembourg\protect\\
29, Avenue J.F Kennedy, L-1855 Luxembourg,\protect\\
e-mail: \{name.surname\}@uni.lu}
\date{}
\maketitle

\begin{abstract}
T-distributed stochastic neighbor embedding (tSNE) is a popular and prize-winning approach for dimensionality reduction and visualizing high-dimensional data. However, tSNE is non-parametric: once visualization is built, tSNE is not designed to incorporate additional data into existing representation. It highly limits the applicability of tSNE to the scenarios where data are added or updated over time (like dashboards or series of data snapshots).

In this paper we propose, analyze and evaluate LION-tSNE (Local Interpolation with Outlier coNtrol) - a novel approach for incorporating new data into tSNE representation. LION-tSNE is based on local interpolation in the vicinity of training data, outlier detection and a special outlier mapping algorithm. We show that LION-tSNE method is robust both to outliers and to new samples from existing clusters. We also discuss multiple possible improvements for special cases.

We compare LION-tSNE to a comprehensive list of possible benchmark approaches that include multiple interpolation techniques, gradient descent for new data, and neural network approximation.
\end{abstract}


\section{Introduction}

	Exploring high-dimensional datasets is a problem that frequently arises in many areas of science, and visualization is an important aspect of data exploration. Throughout the years many approaches were developed to reduce the data dimensionality and to visualize multidimensional data. Notable examples include principal component analysis (PCA) \cite{hotelling_analysis_1933}, Sammon mapping \cite{sammon_nonlinear_1969}, Isomap \cite{tenenbaum_global_2000}, local linear embedding \cite{roweis_nonlinear_2000}, and Chernoff's faces \cite{chernoff_use_1973}. The choice of dimensionality reduction method depends on what are the main characteristics of interest in the data: those can be  inter-point distances \cite{sammon_nonlinear_1969}, outliers \cite{chernoff_use_1973}, linear subspaces \cite{hotelling_analysis_1933}, and more.
	
	T-distributed stochastic neighbor embedding (tSNE) \cite{maaten_visualizing_2008} is a very popular prize-winning \cite{noauthor_t-distributed_2012} algorithm for dimensionality reduction. The main strength of tSNE is visualizing clusters: if the data points are close in high-dimensional space, tSNE aims to keep data close in reduced dimensions as well. Hence, the clusters in original data should be visible after dimensionality is reduced. Throughout the years tSNE has been successfully applied to visualizing genomic data \cite{li_application_2017}, healthcare information \cite{nguyen_m-tsne:_2016}, human gait \cite{che_gait_2015}, power consumption profiles \cite{moran_analysis_2012}, heating systems \cite{dominguez_dimensionality_2015}, tumor subpopulations \cite{abdelmoula_data-driven_2016}, and many other application areas. 
	
	The main limitation of tSNE is that it learns non-parametric mapping: once the data are visualized, adding new data samples to existing visualization is not trivial. The same problem also highly limits the possibility to work with time-varying data: if the data point has changed, it should be visualized at a new location, but tSNE is not designed for finding new data locations within existing visualizations. While there were some attempts to extend tSNE to incorporate new data \cite{maaten_learning_2009, gisbrecht_parametric_2015}, a large amount of possible approaches remains unexplored, and there was no attempt to create a systematic comparative study.
	
	This work addresses the problem of embedding new data into existing tSNE representations. We propose LION-tSNE (Local Interpolation with Outlier coNtrol) - a novel method for incorporating new data samples into existing tSNE representation. We identify that most interpolation and approximation methods have a lack of robustness to newly introduced outliers. Another problem is that when a new data sample is introduced, very distant existing visualized points have disproportionate influence on the mapping of a new point, and often it results in significant performance loss. Those two challenges can be naturally solved by a single approach: use local interpolation/approximation, and invoke a special outlier mapping procedure when necessary. This is the main idea of LION-tSNE, and section \ref{section_lion_tsne} provides detailed description of the algorithm.
	
	This paper is structured as follows. Understanding of tSNE algorithm is a requirement, so section \ref{section_tsne_intro} provides basic overview of tSNE. It also describes the limitations of the original tSNE algorithm. Section \ref{section_related_work} describes related work and defines the goals of the paper. Section \ref{section_lion_tsne} gives a detailed description of the LION-tSNE approach. Section \ref{section_methods} proposes classification and a list of other plausible mapping methods for the tSNE algorithm. Those methods are then used as benchmarks for LION-TSNE in section \ref{section_evaluation}. Section \ref{section_evaluation} provides analysis, evaluation and comparison of LION-tSNE to various interpolation and approximation methods. Section \ref{section_discussion} provides the discussion and summarizes obtained results. Section \ref{section_conclusion} discusses the directions of future work and concludes the paper.

\section{T-Distributed Stochastic Neighbor Embedding} \label{section_tsne_intro}

T-Distibuted Stochastic Neighbor Embedding algorithm was developed by L. van der Maaten and  G. Hinton \cite{maaten_visualizing_2008}. TSNE itself is based on SNE algorithm \cite{hinton_stochastic_2003} by G. Hinton and S. Roweis. TSNE takes multidimensional data as an input and returns the representation of that data in a space with reduced dimensions (so-called embedded space, usually 2-dimensional). TSNE treats distances between all pairs of data samples as a distribution. There are two such distributions: one for distances in the original space and one for distances in the embedded space. TSNE preserves the distances by minimizing Kullback-Leibler divergence (i.e. distance) between those two distributions. Mainly due to the asymmetric nature of Kullback-Leibler divergence, the tSNE algorithm favors keeping close points close. Here we provide only a brief explanation of tSNE algorithm. Please, refer to original articles \cite{maaten_visualizing_2008} and \cite{hinton_stochastic_2003} for more details and explanations.

Consider a dataset $X \in \mathbb{R}^{N\textnormal{x}K}$, which is referred to as the training set. It consists of $N$ data points, each point is $K$-dimensional, and there is a distance metric defined in $K$-dimensional space. The task is to find an embedding $Y \in \mathbb{R}^{N\textnormal{x}d}$, where $d$ is usually 2 or 3. Here we refer to some value in the original high dimensional space as $x$, denoting $x_i$ if we are talking about a sample from the training set $X$. Similarly, we refer to a point in the embedded space as $y$, denoting $y_i$ if it is a point found by tSNE algorithm for training set sample $x_i$. For simplicity we are going to refer to original K-dimensional space as $x$ space, and to reduced dimensionality $d$-dimensional space as $y$ space.  

For the purpose of finding a mapping $f: X \rightarrow Y$, in this paper we assume that all $x_i$ are different. It can be assumed without loss of generality - if $x_i = x_j$ we can just treat them as one data point and remove duplicate row from the training set.

The algorithm accepts one parameter, perplexity, which can be roughly thought as the expected number of neighbors. Proper perplexity is often chosen manually by running tSNE several times and picking best option, but it should be noted that tSNE is fairly robust to different perplexity choices.

\begin{figure*}
	\centering
	\caption{tSNE representation of MNIST dataset: 28x28 grayscale handwritten digits, preprocessed using PCA (first 30 components used). Perplexity = 30.}
	\label{fig_mnist_tsne_original}
	\includegraphics[width=0.7\figurewidth]{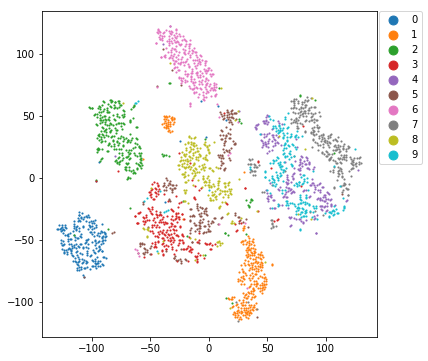}
\end{figure*}

Throughout the paper we will use the MNIST dataset (Modified National Institute of Standards and Technology dataset) as a running example. MNIST is a dataset of 28x28 grayscale handwritten digits \cite{lecun_mnist_nodate}. Among many other use cases, MNIST was used as one of the benchmarks for the original tSNE algorithm \cite{maaten_visualizing_2008}. Figure \ref{fig_mnist_tsne_original} refers to the tSNE algorithm with a perplexity of 30 applied to 2500 random distinct samples of MNIST digits, compressed to first 30 principal components. Here $X$ refers to entire 2500x30 digit dataset, $Y$ refers to 2500x2 visualizations of each point, and $x_i$ and $y_i$ refer to any particular handwritten digit and its visualization respectively. Throughout this paper we are going to deploy new 28x28 grayscale images into the visualization (figure \ref{fig_mnist_tsne_original}) and evaluate whether new images are at the plausible position.

TSNE operates in a following manner. At first, pairwise distances are represented as a following distribution (formula \ref{eq_P_conditional}).

\begin{equation} \label{eq_P_conditional}
p_{i|j} =  \frac{\exp{\frac{\|x_i - x_j\|^2}{2\sigma_i^2}}}
{\sum_{k,i \neq k} \exp{\frac{\|x_i - x_k\|^2}{2\sigma_i^2}} }
\end{equation}

The variances $\sigma_i$ for each point is determined using equation \ref{eq_perplexity_vs_entropy}.

\begin{equation} \label{eq_perplexity_vs_entropy}
Perp(i) =  2^{H(P_i)}
\end{equation}

In formula \ref{eq_perplexity_vs_entropy} $Perp(i)$ is perplexity of the $i$-th point. All perplexities (usually a single value for all the points) are given as input parameters, and $H(P_i)$ is Shannon entropy and can be calculated as follows (equation \ref{eq_shannon_entropy}).

\begin{equation} \label{eq_shannon_entropy}
H(P_i) = \sum_{j,i \neq j} p_{i|j} log_2(p_{i|j})
\end{equation}

Then the variances $sigma_i$ can be calculated using root finding techniques (the authors of tSNE \cite{maaten_visualizing_2008} suggested binary search).

Symmetric joint distribution can then be achieved by symmetrizing $p_{i|j}$ in the following manner (expression \ref{eq_symmetrization})

\begin{equation}\label{eq_symmetrization}
p_{ij} = \frac{p_{i|j} + p_{j|i}}{2N}
\end{equation}

Distances in the embedded space are represented as distributions in quite similar manner, but t-distribution is used instead of Gaussian distribution. Distribution in the embedded space is referred to as $Q$ matrix, where each element $q_{ij}$ is calculated as follows (formula \ref{eq_q_matrix}).

\begin{equation} \label{eq_q_matrix}
q_{i|j} =  \frac{(1 + \|y_i - y_j\|^2)^{-1}}
{\sum_{k,i \neq k} (1 + \|y_i - y_k\|^2)^{-1}}
\end{equation}

The goal is to find vectors $y_i$ for each data point, such that Kullback-Leibler divergence of distributions $P$ and $Q$ (i.e. distance between distributions $P$ and $Q$) is minimal. So, the task is to minimize following function w.r.t. $Y$ (see expression \ref{eq_KL_divergence}).

\begin{equation} \label{eq_KL_divergence}
C(Y) = KL(P\|Q) = \sum_{i,j,i\neq j}p_{ij} log \frac{p_{ij}}{q_{ij}}
\end{equation}

Resulting embedded coordinates $Y$ are calculated by minimizing the function \ref{eq_KL_divergence} using gradient descent, and the gradient can be expressed analytically (formula \ref{eq_KL_gradient}).

\begin{equation} \label{eq_KL_gradient}
\frac{\partial C}{\partial y_i} = 4 \sum_{j,j \neq i}(p_{ij} - q_{ij})(y_i - y_j)(1 + \|y_i - y_j\|^2)^{-1}
\end{equation}

The authors of original tSNE algorithm \cite{maaten_visualizing_2008} suggest several improvements for gradient descent to avoid poor local optima. Those include using momentum and early exaggeration - multiplying $P$ matrix by a factor for first multiple iterations of gradient descent algorithm.

The essential limitation of the tSNE algorithm is that it builds non-parametric mapping. Once the embedding for $X$ is built, there is no straightforward formula or approach to visualize a new sample $x$ in the same visualization. This new $x$ can be either some completely new sample, or an update of an existing sample $x_i$ in the time-varying data scenario. The next section discusses existing attempts to address that limitation and refines the research subject of this paper.

\section{Related Work} \label{section_related_work}

	The idea of incorporating new data into tSNE is a recognized problem. Existing solutions mainly took the form of finding a mapping function $f: X\rightarrow Y$, which accepts a point in $K$-dimensional original space and returns its embedding in 2 or 3 dimensions. Once this mapping is found, the function can be used to incorporate new samples.
	
	L. van der Maaten \cite{maaten_learning_2009}, the author of the original tSNE algorithm, addressed the problem of building $f: X\rightarrow Y$ representation for tSNE. The author proposed using restricted Boltzmann machines to build an approximation of tSNE mapping and compared the results to non-tSNE based visualization approaches. Rather than interpolating or approximating existing tSNE results, the author learned both tSNE representation and mapping together. Parametric tSNE is one of the benchmark approaches to compare against LION-tSNE. The main difference in the approaches is that LION-tSNE aims to represent existing tSNE visualization, while parametric tSNE aims to learn both mapping and tSNE together. Performance comparison is provided in section \ref{section_large_data}.
	
	Gisbrecht et. al \cite{gisbrecht_parametric_2015} proposed kernelized tSNE - parametric tSNE based on normalized Gaussian kernels. This approach is close to RBF interpolation \cite{buhmann_radial_2003} and also has some resemblance to IDW interpolation \cite{shepard_two-dimensional_1968}. Earlier version of the same method was also proposed in \cite{gisbrecht_out--sample_2012}. We use kernelized tSNE method as one of the benchmarks to compare against LION-tSNE in terms of cluster attribution accuracy and outlier robustness. The main difference between LION-tSNE and kernelized tSNE are two main features of LION-tSNE method: its use of local interpolation and its special focus on outlier handling. Extensive performance comparison of LION-tSNE and kernelized tSNE is provided in section \ref{section_evaluation}.
	
	Several related work items also dealt with either the question of adding new data to tSNE or using tSNE for dynamic data, but they did it in another context and solved different, although adjacent, problems. Due to that, those approaches cannot be compared directly with LION-tSNE, but section \ref{section_discussion} discusses, in which practical circumstances can those methods be chosen, and in which practical scenarios LION-tSNE approach should be used. 
	
	Pezotti et al. \cite{pezzotti_approximated_2017} proposed A-tSNE - an improvement over tSNE for progressive visual analytics. The authors used K-Nearest Neighbors (KNN) to approximate distance functions for tSNE calculations, and used degree of approximation to vary the tradeoff between precision and calculation time. Pezotti et al. \cite{pezzotti_approximated_2017} also included procedure for removing and adding point to tSNE based on adding/removing points from KNN neighborhoods.
	
	There were also multiple attempts to visualize time-varying data using tSNE, but avoiding the problem of incorporating new samples. Notable examples include the paper of Rauber et al. \cite{rauber_visualizing_2016}. The authors proposed dynamic tSNE, along with several strategies for calculating $P$ matrices at each time step. Nguen et al. \cite{nguyen_m-tsne:_2016} proposed mTSNE algorithm, where the authors created single visualization for the entire set of time series, using time series distance metrics to produce $P$ matrix.
	
	While several papers proposed to build a $f: X\rightarrow Y$ mapping for the tSNE algorithm, they mostly focus on a single mapping approach, rather than comparing alternatives. A large variety of plausible mapping approaches remain unexplored. There is a lack of classification and a comparative study. It is not explored how different mapping approaches perform under different circumstances, e.g. what happens if an outlier is added. These are the questions our paper addresses.
	
	In this paper we explore, evaluate and compare a comprehensive list of plausible approaches for incorporating new data into tSNE representation. The list goes beyond the approaches suggested in related work, and we use it as a benchmark to compare with our newly proposed algorithm. We show that while some approaches work well in the vicinity of training data, and a few approaches can handle outliers, there is no approach that achieves both goals, and the performance on those two tasks is often inversely related. We propose, analyze and evaluate LION-tSNE - a method that can handle well those two challenges at once.
		
	The next section introduces LION-tSNE (Local Interpolation with Outlier coNtrol) - an algorithm for incorporating new data into tSNE representation.

\section{LION-tSNE Algorithm} \label{section_lion_tsne}

	LION-tSNE algorithm is designed to have two important features:
	\begin{description}
		\item[Outlier robustness.] Many existing methods work well if a new sample $x$ is in the vicinity of an existing training points $x_i$. However, those methods struggle when presented with outliers (see section \ref{section_evaluation} for details). Outlier detection and special outlier handling algorithm can help to alleviate that problem.
		\item[Locality.] As will be shown in section \ref{section_evaluation}, for a new sample $x$ even very distant training points can significantly influence its embedding $y$. And the influence of distant points can very significantly reduce the performance. This problem is applicable to all benchmark methods, while being especially visible in inverse distance weighting interpolation. A solution to this problem is to ensure locality - embedding of new sample $x$ should depend only on training samples $x_i$ within a certain radius.
	\end{description}

	Outlier detection and locality can naturally combine into a single approach: we can use local interpolation when a new sample $x$ has neighbors in a certain radius $r_x$, and we can invoke outlier placement procedure otherwise. The radius $r$ is referred to as neighbor radius and it is a parameter of the algorithm. On a very general level LION-tSNE approach can be summarized in algorithm \ref{alg_lion_basic}.
	
	\begin{algorithm}
	\caption{LION-tSNE - General Approach \label{alg_lion_basic}}
	\begin{algorithmic}[1]
		\Function{LION-tSNE}{$x$, $p$, $r_x$, $r_close$, $X_{train}$, $Y_{train}$}
		\State $neighbor\_indices = select\_neighbors\_in\_radius(x, X_{train}, r_x)$
		\State $X_{neighb} = X_{train}[neighbor\_indices]$
		\State $Y_{neighb} = Y_{train}[neighbor\_indices]$		
		\If{$len(neighbor\_indices) > 1$}
			\State $y = local\_IDW\_interpolation(X_{neighb},	Y_{neighb})$		
		\ElsIf{$len(neighbor\_indices) == 1$}
			\State $y = single\_neighbor\_placement(X_{neighb},	Y_{neighb})$		
		\Else
			\State $y = outlier\_placement()$
		\EndIf		
		\Return $y$
		\EndFunction
	\end{algorithmic}
	\end{algorithm} 	
	
	For local interpolation here we use inverse distance weighting (IDW) \cite{shepard_two-dimensional_1968}. The value for new point $x$ is determined as weighted sum of values $y_i$, where weight is proportional to inverse distances. It can be formalized as follows (formula \ref{eq_IDW})
	
	\begin{equation} \label{eq_IDW}
	F(x) = \sum_{\|x-x_i\| \le r} w_i(x) y_i \textnormal{, where } w_i(x) = \frac{\|x-x_i\|^{-p}}{\sum_{\|x-x_j\| \le r} \|x-x_j\|^{-p}}
	\end{equation}
	
	When the point $x \rightarrow x_i$, the inverse distance $\|x - x_i\|^{-1} \rightarrow \infty $, the corresponding weight $w_i(x) \rightarrow 1$ (and all other weights $\rightarrow 0$ due to normalization) and $F(x) \rightarrow y_i$. If $x = x_i$ the mapping $y$ is forced to $y_i$. The power $p$ is the parameter of the algorithm. Note that for LION-tSNE we use local version of IDW interpolation - algorithm considers only those training samples where $\|x-x_i\| \le r$.
	
	IDW interpolation is relatively fast and produces high accuracy in the vicinity of training samples (see section \ref{section_evaluation} for accuracy estimates and complexity analysis). As an interpolation method, it ensures consistency: mapping for $x = x_i$ will produce at $y = y_i$. Also it is capable of producing multidimensional output $y$.
	
	IDW requires to carefully choose the algorithm parameter $p$. Also IDW has serious shortcomings that will be discussed in sections \ref{section_methods} and \ref{section_evaluation}, but those shortcomings mainly appear when sample $x$ is far away from the training samples, and LION-tSNE uses IDW approach only locally. RBF interpolation \cite{buhmann_radial_2003} and some approximation methods are plausible substitutes of IDW in LION-tSNE.
	
	There is a special case when for some new sample $x$ there is only one neighbor in $r_x$ - meaningful interpolation requires at least two neighbor points. In that case behavior depends on whether that single neighbor $x_i$ is an outlier itself.
	\begin{itemize}
	\item If $x_i$ is an outlier too, it will be more meaningful from a visualization perspective to put embedding $y$ of new sample $x$ in the vicinity of $y_i$ - we have several outlier cases that are close to each other. 
	\item In case if $x_i$ is not an outlier, it is recommended to denote $x$ as an outlier and use the outlier placement procedure - $x$ does not really belong to the same cluster as $x_i$ and should not be treated as such.
	\end{itemize}
	
	By \textit{placing close} to some $y$ or \textit{placing in the vicinity} of some $y$ from now on we mean placing at some random position at a distance not exceeding $r_{close}$, where $r_{close}$ can be set to some low percentile of distribution of nearest neighbor distances in $y$ space (like 10\% or 20\%). That way all points \textit{placed close} together will form a cluster as dense as the clusters in current visualization. 
	
	\subsection{Outlier Placement}\label{section_outlier_placement}
	
	\begin{figure*}
	\centering
	\caption{Outlier placement: example. Radius $r_y$ was set at the largest nearest neighbor distance in $y$ space, without multiplying by coefficient or adding safety margin like $r_{close}$. Outliers were generated randomly and mapped by LION-tSNE all at once.}
	\label{fig_outlier_placement_example}
	\includegraphics[width=0.7\figurewidth]{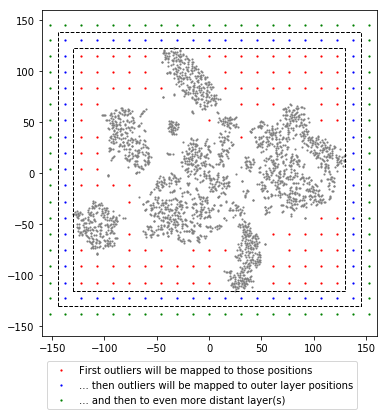}
	\end{figure*}		
	
	The main goal of the outlier placement is following: if a new sample $x$ is an outlier, its mapping $y$ should be perceived as an outlier in the visualization. In order to place an outlier, we need to find a position for $y$, such that there are no neighbors in a certain radius $r_y$. Using something like $y \gg y_{max}$ for each dimension of $y$ is not an option - placing new sample too far from existing values will complicate the visualization.
	
	The radius $r_y$ is a parameter of the algorithm. Too large $r_y$ can put outlier points too far from the existing plot area and reduce the readability of the visualization, while too small $r_y$ can make clusters and outliers indistinguishable. Radius $r_y$ can be based on $r_{yNN}$ - some percentile of the distribution of the nearest neighbor distances in $y$ space, e.g. 95th, or 99th percentile depending on what percentage of original training points we assume to be outliers; or $r_{yNN}$ can be set to maximum distance between nearest neighbors in $y$ space. For better visibility $r_{yNN}$ can be further multiplied by a coefficient $k \approx 2$. Also it is recommended to add some safety margin to $r_y$ - if several outliers are close to each other, they will be placed at a distance of $r_{close}$, and distance between one of them and different outlier can become lower than the chosen radius. In addition, one of the embedded points could get close to outlier if it is placed close to its only neighbor. Therefore, the recommended choice is $r_y \ge k*r_{yNN} + r_{close}$.  However, $r_y$ can be chosen relatively arbitrary: its main purpose is just the clarity of visualization, and as long as outliers and clusters are clearly distinguishable, any choice of $r_y$ is acceptable.
	
	The goal of outlier placement algorithm is to find positions in $y$ space, such that there are no training data $y_i$ within $r_y$ radius. Additional challenge is that multiple different outliers can be introduced at once, and LION-tSNE should find positions for all of them. Multiple outliers should be at least $r_y$ away both from training samples and from each other. Also if possible the size of the plot should stay the same - new data should not be mapped to the area outside of $[min(y_i), max(y_i)]$ (for each dimension of $y$), unless it is necessary.
	
	The main idea of outlier placement algorithm is following: in order to find possible outlier mapping positions in $y$ space it is sufficient to find set of square-shaped cells with the side $2r_y$, such that there are no training data $y_i$ within each cell and those cells do not overlap. If $y$ space is not 2-dimensional, cubes or hypercubes are used instead of square-shaped cells. If an outlier is mapped to the center of that cell, it is guaranteed that no training samples are within $r_y$ radius. And if each outlier mapping occupies a different cell, then outlier mappings are at least at $r_y$ distance from each other as well. In practice the size of each cell along each dimension is slightly increased in order to fit exact number of equally sized cells between $[min(y_i), max(y_i)]$. The positions of those cells can be found using the approach described below in the algorithm \ref{alg_outlier_positioning}.
	
	\begin{algorithm}
	\caption{Computing Possible Outlier Positions\label{alg_outlier_positioning}}
	\begin{algorithmic}[1]
		\Function{PrecomputeOutlierPositions}{$r_y$, $Y_{train}$}
		\State // Adjust cell size per dimension to have exact number of cells
		\State $n\_cells\_per\_dimension = \lfloor\frac{max(Y_{train})-min(Y_{train})}{r_y}\rfloor$
		\State $r_{adj} = (max(Y_{train})-min(Y_{train}))/n\_cells\_per\_dimension$
		\State // Find how many cells can we fit per each dimension
		\State // Generate all possible cell coordinates
		\State $all\_cells = all\_combinations(n\_cells\_per\_dimension)$
		\State $free\_cells = list()$		
		\ForAll{$cell \in all\_cells$}
		\State // Find cell boundaries
		\ForAll{$i \in cell.dimensions$}
			\State $cell\_bounds.min = min(Y_{train}) + cell_i*r_{adj,i}$
			\State $cell\_bounds.max = min(Y_{train}) + (cell_i+1)*r_{adj,i}$
		\EndFor		
		\If {$no\_samples\_in\_cell(cell\_bounds, Y_{train})$}
			\State $free\_cells.append(cell)$
		\EndIf
		\EndFor
		\Return $free\_cells.centers()$
		\EndFunction		
	\end{algorithmic}
	\end{algorithm} 	
	
	The case when several outliers need to be placed at once is more complicated. In that case not only the distance to existing training points, but also the distance between the outliers themselves needs to be maintained. If some outliers are distinct from existing training points, but have distance $\le r$ from each other, it is recommended to use outlier placement procedure only for one of those outliers, and place all other ones at some close distance in $y$ space to that sample. Again, close distance can be defined as some low percentile of distribution of nearest neighbors distances in $y$ space (e.g. 10th or 20th percentile). For all remaining distinct outliers the positions precomputed by algorithm \ref{alg_outlier_positioning} can be chosen at random. The work of the algorithm is illustrated on MNIST dataset in figure \ref{fig_outlier_placement_example}.	
	
	To summarize, the positions for outliers are precomputed in advance. Then, if necessary, position for outlier placement can be taken from the pool of available positions in constant time. If there are no more precomputed available positions, new positions outside current plotting area can be precomputed as shown in figure \ref{fig_outlier_placement_example}.
	
	While this concludes the general description of LION-tSNE, the algorithm still requires setting several parameters. The next section summarizes methods for setting those algorithm parameters and proposes an approach for selecting power parameter $p$.
	
	\subsection{Selecting Power Parameter}\label{section_selecting_power_parameter}
	
	\begin{figure*}
		\centering
		\caption{Leave-one-out cross-validation results.}
		\label{fig_power_toy_example}
		\includegraphics[width=0.7\figurewidth]{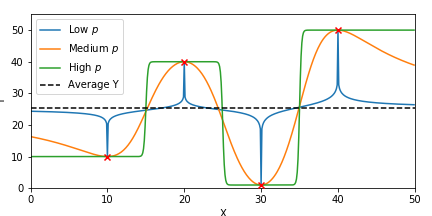}
	\end{figure*}	
	
	The LION-tSNE algorithm has 3 parameters: $r_x$, $r_y$ and power $p$. The parameter $r_x$ can be chosen based on the percentile of nearest neighbor distances: e.g. if we assume 1\% of initial data to be outliers, $r_x$ can be set to 99th percentile of nearest neighbor distances in $x$ space. The parameter $r_y$ can be chosen relatively arbitrary, and some heuristics for selection are proposed in section \ref{section_outlier_placement}. However, the selection of the remaining parameter $p$ is not that straightforward.
	
	Proper choice of power parameter $p$ is paramount for IDW interpolation. In order to understand the effect of parameter $p$ consider figure \ref{fig_power_toy_example}. It shows a simple example of IDW interpolation for one-dimensional $x$ and one-dimensional $y$, but the same effect is applicable to multidimensional IDW interpolation as well. In the example interpolated (x,y) pairs are (10,10),(20,40),(30,1), and (40,50), and the powers 0.2, 2 and 20 were tested. For low power even when $x$ looks relatively close to $x_i$, the weight distribution is still close to uniform, and predicted value is around the center: $y \approx mean(y_i)$ (unless $x$ is really very close to some $x_i$). When the distance $\|x-x_i\|$ is low and the power is high, the weight $w_i(x)$ becomes dominating if $x$ is even slightly closer to $x_i$ than to any other training sample. As a result, $y \approx y_i$, where $i = argmin\|x-x_j\|$. Too small and too large parameter $p$ can be viewed as different forms of overfitting - in both cases training samples are remembered, rather than generalized. Still properly chosen power $p$ produces good interpolation. To summarize, the parameter $p$ determines the generalization capabilities, and the goal is to evaluate those capabilities using only the training data.
	
	\begin{algorithm}
		\caption{Metrics for parameter $p$\label{alg_cross_validation}}
		\begin{algorithmic}[1]
			\Function{ValidationMetrics}{$p$, $r_x$, $X_{train}$, $Y_{train}$}
				\ForAll{$x_i,y_i \in X_{train},Y_{train}$}
					\State $neighbor\_indices = select\_neighbors\_in\_radius(X_{train}, x_i, r_x)$
					\State $neighbor\_indices.exclude(i)$
					\State $X_{neighb} = X_{train}[neighbor\_indices]$
					\State $Y_{neighb} = Y_{train}[neighbor\_indices]$					
					\State $y_{estimate} = local\_IDW\_interpolation(X_{neighb},	Y_{neighb})$
					\State $sum\_square\_error += \|y_{estimate} - y_i\|$
				\EndFor
				\State $mean\_square\_error = sum\_square\_error / length(X_{train})$
				
			\Return $mean\_square\_error$
			\EndFunction
		\end{algorithmic}
	\end{algorithm} 	

	\begin{figure*}
	\centering
	\caption{MNIST: power parameter influence on IDW interpolation.}
	\label{fig_power_cross_validation}
	\includegraphics[width=0.7\figurewidth]{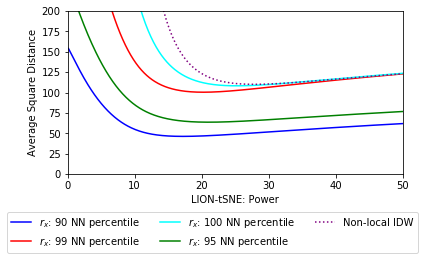}
	\end{figure*}	
	
	In order to test generalization capabilities we propose a variation of leave-one-out cross-validation, that works in a manner described in algorithm \ref{alg_cross_validation}. The algorithm iterates over training samples, and for each sample it performs local IDW interpolation using neighbors of the sample, but not the sample itself. Then we estimate the value at $x_i$ and compare it with expected $y_i$. The average square distance between estimated $y_i$ and real $y_i$ is the final metrics returned by the algorithm. An assumption is that if only one sample is left out at a time, the optimal power $p$ should not shift much, and power $p$ that performs best on cross-validation should effectively remain good choice. Results for MNIST training data are presented in figure \ref{fig_power_cross_validation}. If power is too low, the interpolation function effectively returns mean $y$ almost all the time, and the error is high. In case if power is too high, interpolation effectively returns nearest training sample $y_i$, and it also increases error metrics. However, there is a clear minimum in the metric function. Note that the average mean square error looks relatively high, but the value itself is not the metric of how good the algorithm works, it provides ground only for comparison. Evaluation metrics will be further discussed in section \ref{section_evaluation}.  Also later in section \ref{section_evaluation} we investigate further the influence of power parameter and confirm that the proposed metrics really allows choosing proper power $p$. However, this metric is a heuristic, not an exact criterion. As will be shown in section \ref{section_evaluation}, for low $r_x$ choosing slightly higher power values can produce slightly better results. Also both figure \ref{fig_power_cross_validation} and results from section \ref{section_evaluation} show that LION-tSNE is more robust to choosing too high values of $p$, rather than to choosing too low values of $p$.
	
	Selected $p$ are summarized in table \ref{tab_power_selection}. For each $r_x$ selection we set power $p$ at the minimum of cross validation metrics.
	
	\begin{table*} \caption{Power parameter selection}  \label{tab_power_selection}
	\begin{tabular}{| M{0.23\textwidth} | M{0.10\textwidth}| M{0.10\textwidth}| M{0.10\textwidth}| M{0.10\textwidth}| M{0.15\textwidth} |}
		\hline
		\textbf{$r_x$ percentile} &90 & 95 & 99 & 100 &  Non-local IDW
		\\ \hline
		\textbf{Selected $p$} &17.2 &21.3 &20.7 &25.5 &27.9
		\\ \hline
	\end{tabular}
	\end{table*}
	
	This concludes the description of all aspects of the LION-tSNE approach. Next section introduces several benchmark methods, to which LION-tSNE will be compared.

\section{Benchmark Methods} \label{section_methods}

This section provides the description of benchmark algorithms for incorporating new data into existing tSNE visualization. There are several major approaches to that problem.

\textbf{Repeated gradient descent.} A possible approach is to add new sample $x$ to training matrix $X$, and then continue running tSNE - repeat gradient descent again, until new minimum of cost function is reached. The approach can be summarized as follows.

\begin{enumerate}
	\item Add new sample $x$ to the matrix of training samples $X$. Let it be matrix $X'$ that now contains $N+1$ rows, new $x$ corresponding to the last row.
	\item Recalculate new $P'$ matrix for $X'$. The option is to keep old values of $\sigma_i$ or calculate them again using the new data.
	\item Run gradient descent to find new  embedding $y$ by minimizing KL divergence (formula \ref{eq_KL_divergence}). In order for original $y_i$ points to stay in place, the gradients $\sfrac {\partial{KL(P\|Q)}}{\partial{y_i}}$ should be forced to 0 for $i=1\dots N$. The choice is whether to use early exaggeration for new gradient descent or not. Another choice is how to initialize starting $y$ for gradient descent. 
\end{enumerate}

\textbf{Building embedding function} is another possible approach to incorporating new data into tSNE mapping. After tSNE run we have a set of training samples $x_i$ and corresponding $y_i$. The task is to find a function $f: x \to y$, which then will be used to produce mapping for new samples. It can be viewed as an interpolation or approximation task. Interpolation ensures consistency: for each training point $x_i$ the produced mapping will be exactly $y_i$. However, approximation might generalize better. Approximation function does not fit all the data points exactly, and at first glance consistency will be lost. However, there is a way to avoid the loss of consistency: we can build approximation function right after the first application of tSNE, and from that point and on work with approximated embedding only.

As one of the benchmark options we use inverse distance weighting interpolation, which was discussed in section \ref{section_lion_tsne}. Note that for benchmark we use non-local IDW interpolation - all training samples participate in the weighted sum in formula \ref{eq_IDW}.

Another option that we explore is radial basis functions (RBF) interpolation. Buhmann \cite{buhmann_radial_2003} provides a thorough introduction to radial basis functions. Interpolation using RBFs is performed in a following manner (formula \ref{eq_RBF_main}).

\begin{equation} \label{eq_RBF_main}
 F(x) = \sum_{i=1}^{N} \lambda_i \phi (\|x-x_i\|)
\end{equation}

In formula \ref{eq_RBF_main} $\phi (\|x-x_i\|)$ is a radial function, i.e. it accepts multidimensional input, but depends only on distance between the input argument and some reference point \cite{buhmann_radial_2003}.

The interpolation should pass through all the points in a training set, therefore, the equations \ref{eq_RBF_fit} should be satisfied.

\begin{equation} \label{eq_RBF_fit}
y_j = F(x_j) = \sum_{i=1}^{N} \lambda_i \phi (\|x_j-x_i\|) \textnormal{, for j=1\dots N}
\end{equation}

Coefficients $\lambda_i$ can be found by solving the system of linear equations (SLE) \ref{eq_RBF_fit}, and under some conditions invertibility of SLE matrix is guaranteed \cite{buhmann_radial_2003}.

\begin{figure*}
	\centering
	\caption{Neural network approximation: grid search results. Value represents average square distance in  on a validation set of 500 MNIST digits.}
	\label{fig_nn_grid_search}
	\includegraphics[width=\figurewidth]{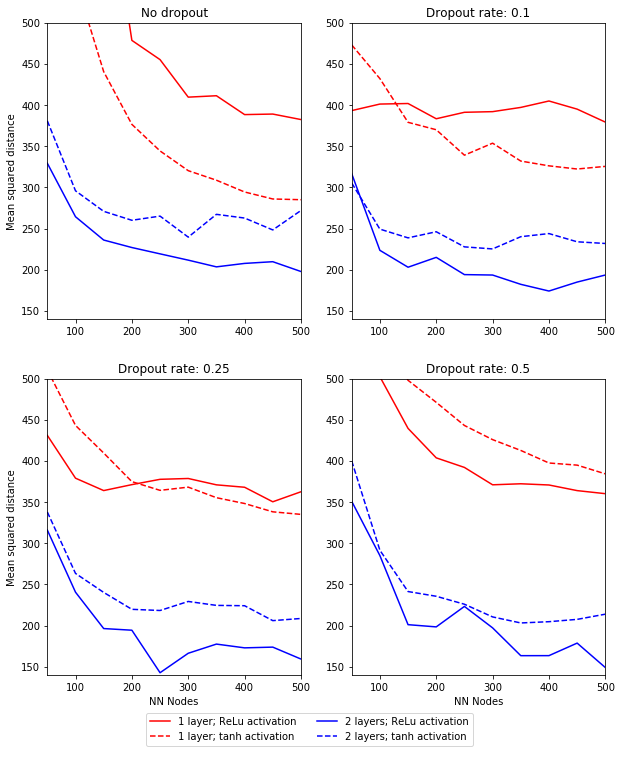}
\end{figure*}

Along with interpolation, we also use approximation for benchmark. Approximating tSNE result requires multivatiate input and mutlivariate output. For that purpose neural networks \cite{haykin_neural_2009} are a natural choice. Applying neural networks for approximating tSNE visulization can be a topic of its own research. Here we compare the performance of LION-tSNE against several benchmark neural network configurations. In order to determine the best neural network configuration we performed the following procedure. Original tSNE run resulted in 2500 $(x_i, y_i)$ mapped pairs. They were split in 80/20 proportion into neural network training set (2000 samples) and neural network validation set (500 samples). Neural network was trained using the 2000-length training set in batch mode for 5000 epochs. The optimization criterion was average square distance between predicted $y$ and corresponding point $y_i$ from validation set. Each combination of following neural network parameters was tested. 
\begin{itemize}
	\item We tested neural networks with 1 and 2 hidden layers. Quick check showed that just adding the next layer does not improve performance much.
	\item The number of nodes in a layer was iterated from 50 to 500 with a step of 50. According to quick check, greater number of nodes did not improve the performance significantly. For benchmark we used the same number of nodes in all layers.
	\item For activation function we tested rectified linear units (ReLu, see \cite{nair_rectified_2010}) $ReLu(x) = max(0,x)$ and hyperbolic tangent $tanh(x) = \frac{e^x - e^{-x}}{e^x + e^{-x}}$.
	\item In order to avoid overfitting, we used dropout regularization \cite{srivastava_dropout:_2014}. Dropout rates of 0.1, 0.25 and 0.5 were tested. An option without regularization was also tested. 
\end{itemize}
The results of grid search are presented in figure \ref{fig_nn_grid_search}. It should be noted that all benchmark configurations have signs of severe overfit: validation set error is significant (consider distance of 10 in figure \ref{fig_mnist_tsne_original}), especially in comparison with training set error (below 0.1 for most configurations). Dropout regularization moderately helps. Approximating tSNE results with neural networks can be a topic of its own research, and further efforts in this area are out of scope of this paper. 

For benchmark we picked 3 parameter sets that performed well on validation.
\begin{description}
	\item [2 hidden layers, 250 nodes per layer, ReLu activation, dropout regularization with rate of 0.25.] This approach performed had the least validation error.
	\item [2 hidden layers, 500 nodes per layer, ReLu activation, dropout regularization with rate of 0.5.] This approach performed had second best validation error.
	\item [1 hidden layers, 500 nodes; tanh activation unit; no regularization.] It showed the best validation performance among one hidden layer models. Though it is likely not the best model choice, it will be interesting for benchmark comparison.
\end{description}
Chosen neural networks were retrained using all 2500 samples of the dataset. Although the parameter values are specific to this particular dataset, the same procedure can be used to determine proper parameters for any dataset.

For benchmark we intentionally used most generic neural network configurations, suitable for any kind of dataset. More specialized solutions can be chosen for special cases. For example, if input $x$ is an image, convolutional neural networks \cite{krizhevsky_imagenet_2012} might be fitting choice.

In addition, we decided to compare LION-tSNE to several methods described in related work. One of them is kernelized tSNE, developed by Gisbrecht et. al \cite{gisbrecht_parametric_2015}. The mapping is interpolated using function \ref{eq_kernelized_tsne}. The parameters $\sigma_i$ are determined as a distance from $x_i$ to nearest neighbor $x_j$, then multiplied by some coefficient $K$, which is a parameter of the algorithm. The coefficients $a_i$ can be calculated by plugging  training data into formula \ref{eq_kernelized_tsne}, getting system of linear equations similar to \ref{eq_RBF_fit} and solving it. We use $K$ = 0.05, 0.25 and 0.50 for benchmark tests, but the effect of parameter $K$ on performance is investigated extensively in section \ref{section_evaluation}.

\begin{equation}\label{eq_kernelized_tsne}
	F(x) = \sum w_i(x) a_i \textnormal{, where } w_i(x) = \frac{exp(\frac{-\|x-x_i\|^2}{2\sigma_i^2})}{\sum_j exp(\frac{-\|x-x_j\|^2}{2\sigma_j^2})}
\end{equation}

Also we compare LION-tSNE with parametric tSNE approach described by van der Maaten \cite{maaten_learning_2009}. In that approach tSNE representation is trained together with restricted Boltzmann machines (RBM) model, which is then used for $x \to y$ mapping. RBM are undirected multilayer models used, among other applications, as the earliest deep learning models (see \cite{hinton_fast_2006}).

To summarize, we compare performance of LION-tSNE to the following benchmark methods.
\begin{itemize}
	\item Repeated gradient descent 
	\item RBF interpolation using a range of popular kernels:
	\begin{itemize}
		\item Multiquadric $ \phi (\|r\|) = \sqrt[2]{(\sfrac{r}{\epsilon})^2 + 1}$
		\item Gaussian: $ \phi (\|r\|) = exp(-(\sfrac{r}{\epsilon})^2)$
		\item Inverse multiquadric: $ \phi (\|r\|) = \frac{1}{\sqrt[2]{(\sfrac{r}{\epsilon})^2 + 1}}$
		\item Linear: $ \phi (\|r\|) = r$
		\item Cubic: $ \phi (\|r\|) = r^3$
		\item Quintic: $ \phi (\|r\|) = r^5$
		\item Thin plate spline:  $ \phi (\|r\|) = r^2log(r)$
	\end{itemize}
	For all of the RBF methods parameter$\epsilon$ was set to average distance between $x_i$.
	\item IDW interpolations with powers 1, 10, 20, 27.9 and 40. The power 27.9 was selected as a minimum of metrics described in section \ref{section_selecting_power_parameter}.
	\item Neural networks with 3 parameter combinations described above.
	\item Kernelized tSNE as described by Gisbrecht et. al \cite{gisbrecht_parametric_2015} with $K$ equal to 0.05, 0.25 and 0.50.
	\item Parametric tSNE as described by van der Maaten \cite{maaten_learning_2009}, using 60000 samples of non-processed 784-dimensional tSNE and 3 hidden layers of restricted Boltzmann machines of 500, 500 and 200 nodes per layer respectively.
	\item LION-tSNE with $radius$ set to 90th, 95th and 99th and 100th percentile of nearest neighbors distance distribution. Power was chosen according to the heuristics described in section \ref{section_selecting_power_parameter}, but the effect of power parameter on performance is also investigated.
\end{itemize}

Next section proposes evaluation criteria, introduces the tests and evaluates LION-tSNE performance against all the benchmarks described above.

\section{Evaluation} \label{section_evaluation}

In order to be useful, mapping method should preserve local structure of the data, just like original tSNE representation did. There are two aspects of preserving structure when adding new data:
\begin{itemize}
	\item If a new data point $x$ belongs to some structure in the original space, and this structure is successfully represented in tSNE, then the point $x$ should belong to the embedding of that structure. For example, if new handwritten digit is added to visualization of MNIST dataset, its embedding should belong to the cluster of the same digits. We will show that most interpolation and approximation methods handle this problem relatively well.
	\item If a new data point $x$ does not belong to any structure, its embedding should not belong to any structure in $y$ space as well. For example, if we attempt to add data point with noise to visualization of MNIST dataset, embedding of new point should not belong to any clusters of handwrtten digits. We show that current embedding methods cannot handle this requirement, and propose a new approach to finding $f: X \to Y$ mapping in presence of outliers.
\end{itemize}

\subsection{Cluster attribution test} \label{section_cluster_attribution_test}

\begin{figure*}
	\centering
	\caption{Test set: first 10 test cases out of 1000}
	\subfloat[Original image. Comparing to closest training sample.\label{fig_nearest_unchosen_neighbors}]{
		\includegraphics[width=\figurewidth]{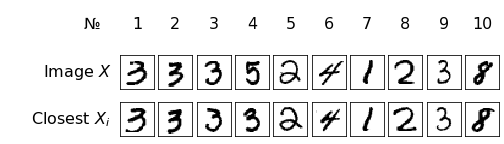}
	}
	\hfill
	\subfloat[Positions of closest training sample. Numbers match figure \ref{fig_nearest_unchosen_neighbors}\label{fig_examples_positions}]{
		\includegraphics[width=\figurewidth]{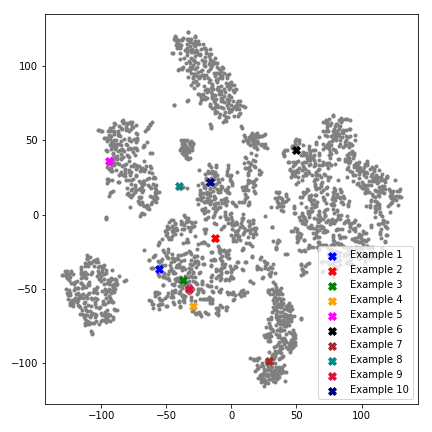}
	}
\end{figure*}

One of the requirements for mapping method is following: if a new sample belongs to some local structure in $x$, it should belong to the same local structure in $y$. For example, in MNIST dataset, if we pick typically written digit "1", and there are clusters of "1" digits in tSNE representation, we expect the new sample to be mapped to a cluster of "1" digits. This property can be tested and evaluated as follows. For MNIST dataset we pick 1000 samples outside of the chosen 2500 training set. In order to make sure that those samples firmly belong to some local structure, we pick only those samples that have a very close neighbor in a training set. In particular, we start from random training set item $x_i$ and pick a new MNIST sample $x$ (outside of the training set) such that $\exists i \forall j, \|x-x_i\| < \|x_j-x_i\|$ - any other training sample $x_j$ is further from $x_i$ than the new sample $x$. Then we generate embedding $y$ for $x$ using various methods, and compare the results. We expect $y$ to belong to the same cluster as $y_i$, or at least belong to another cluster corresponding to the same class. Figure \ref{fig_nearest_unchosen_neighbors} depicts first 10 test images and corresponding $x_i$ from training set. Note that nominal closest neighbor can be of different class (see example 4) due to PCA preprocessing and due to imperfections of distance metrics. Still we expect the mapping method to attribute the test case properly and put in among the neighbors of the proper class.

This test can be evaluated by following criteria:

\textbf{Nearest neighbor accuracy.} New sample should be mapped firmly among the samples of the same class. In order to evaluate that, for each new embedding $x \to y$ we pick K (here we chose $K=10$) nearest neighbors of $y$, and see what percentage of those neighbors is of the same class as the new sample. We compare accuracy of each method to training accuracy that is calcualted as follows. Each test case is designed to be close to some training sample, and even for those training samples 10 nearest neighbors will not always be of the same class. So, even if embedding $y$ is very close to $y_i$, accuracy will not be 100\%. For each test case we estimate 10 nearest neighbors accuracy of the closest training sample. We count it is a practicall upper bound of achievable accuracy, although in theory it can be exceeded (e.g. by chance or by mapping to different cluster of the same class).

\textbf{Nearest neighbor distance.} Test cases are designed to be close to existing training examples. In this test we do not expect new $y$ to be an outlier. Outliers are far from other points, so it can be evaluated by measuring distance to nearest neighbor $y_j$ in the training set. For each training sample we obtain the distance to nearest neighbor and treat those distances as a distribution $D_{NN}$. Then we obtain distance to nearest neighbor for $y$ and determine at what percentile of distribution $D_{NN}$ is that distance. Test cases should be firmly within clusters, so values of around 50\% or below are expected. High values (like 90\% or above) indicate that mapping method produced outliers, rather than put the test samples in the clusters.

\textbf{Kullback-Leibler divergence} between new $P$ and $Q$ distribution matrices (both $(N+1)$x$(N+1)$ - with new test sample incorporated) shows how well data are clustered and how well mapping algorithm keeps original cost function of tSNE. KL divergence can be compared between the methods and also to original KL divergence of the first tSNE run.

Cluster attribution test results are presented in table \ref{tab_cluster_methods_comparison}.

\begin{table*} \caption{Cluster attribution test: methods comparison}  \label{tab_cluster_methods_comparison}
	\begin{tabular}{| m{0.42\textwidth} | m{0.13\textwidth} | m{0.15\textwidth} | m{0.15\textwidth} |}
		\hline
		\textbf{Method}
		& \textbf{Accuracy}
		& \textbf{Distance Percentile}
		& \textbf{KL Divergence}
		\\ \hline
		\textbf{Baseline} & 87.59\% & - & 1.11688
		\\ \hline
		RBF - Multiquadric & 73.84\% & 53.548 & 1.11909
		\\ \hline
		RBF - Gaussian & 73.78\% & 55.081 & 1.11907
		\\ \hline
		RBF - Inverse Multiquadric & 74.30\% & 53.226 & 1.11899
		\\ \hline
		RBF - Linear & 70.93\% & 52.659 & 1.11908
		\\ \hline
		RBF - Cubic & 73.49\% & 53.974 & 1.11909
		\\ \hline
		RBF - Quintic & 67.44\% & 61.672 & 1.11960
		\\ \hline
		RBF - Thin Plate & 72.83\% & 53.141 & 1.11903
		\\ \hline
		IDW - Power 1 & 8.92\% & 65.853 & 1.12220
		\\ \hline
		IDW - Power 10 & 83.73\% & 25.170 & 1.11819
		\\ \hline
		IDW - Power 20 & 87.88\% & 4.845 & 1.11789
		\\ \hline
		IDW - Power 27.9 & 87.95\% & 2.782 & 1.11788
		\\ \hline
		IDW - Power 40 & 87.96\% & 1.905 & 1.11787
		\\ \hline
		GD - Closest $Y_{init}$ & 87.81\% & 10.971 & 1.11780
		\\ \hline
		GD - Random $Y_{init}$ & 33.35\% & 76.420 & 1.12125
		\\ \hline
		GD - Closest $Y_{init}$; new $\sigma$ & 87.81\% & 15.117 & 1.11780
		\\ \hline
		GD - Random $Y_{init}$; new $\sigma$ & 32.51\% & 79.012 & 1.12135
		\\ \hline
		GD - Closest $Y_{init}$; EE & 87.84\% & 10.880 & 1.11780
		\\ \hline
		GD - Random $Y_{init}$; EE & 48.66\% & 56.197 & 1.12019
		\\ \hline
		GD - Closest $Y_{init}$; new $\sigma$; EE & 87.82\% & 15.363 & 1.11780
		\\ \hline
		GD - Random $Y_{init}$; new $\sigma$; EE & 47.81\% & 60.648 & 1.12028
		\\ \hline
		NN - 2L; 250N; ReLu; D0.25 & 78.33\% & 43.473 & 1.37084
		\\ \hline
		NN - 2L; 500N; ReLu; D0.5 & 78.71\% & 42.445 & 1.40617
		\\ \hline
		NN - 1L; 500N; tanh & 72.27\% & 55.941 & 1.13377
		\\ \hline
		Kernelized; K=0.05 & 87.82\% & 0.210 & 1.11788
		\\ \hline
		Kernelized; K=0.25 & 86.47\% & 10.574 & 1.11792
		\\ \hline
		Kernelized; K=0.50 & 73.08\% & 51.203 & 1.11892
		\\ \hline
		\textbf{LION} -  $r_x$ at 90th perc.; $p$=17.2 & \textbf{86.42\%} & \textbf{7.352} & \textbf{1.11797}
		\\ \hline
		\textbf{LION} -  $r_x$ at 95th perc.; $p$=21.0 & \textbf{87.18\%} & \textbf{4.993} & \textbf{1.11793}
		\\ \hline
		\textbf{LION} -  $r_x$ at 99th perc.; $p$=20.2 & \textbf{87.81\%} & \textbf{4.580} & \textbf{1.11790}
		\\ \hline
		\textbf{LION} -  $r_x$ at 100th perc.; $p$=25.2 & \textbf{87.87\%} & \textbf{3.209} & \textbf{1.11788}
		\\ \hline
		
	\end{tabular}
\end{table*}

\begin{figure*}
	\centering
	\caption{Cluster attribution test: RBF interpolation. First 10 test cases.}
	\label{fig_cluster_rbf_all}
	\subfloat[Positioning. Closest training set image connected to embedding results for test image. \label{subfig_cluster_RBF_pos}]{
		\includegraphics[width=\figurewidth]{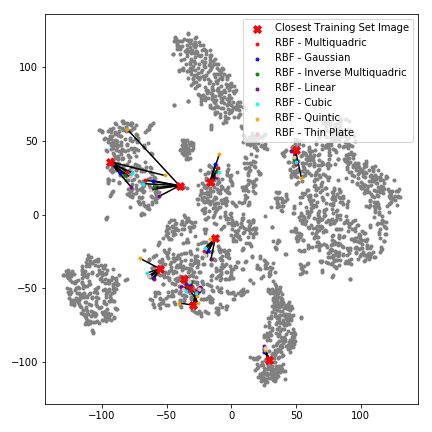}
	}
	\hfill
	\subfloat[Average 10 nearest neighbors\label{subfig_cluster_RBF_avg}]{
		\includegraphics[width=\figurewidth]{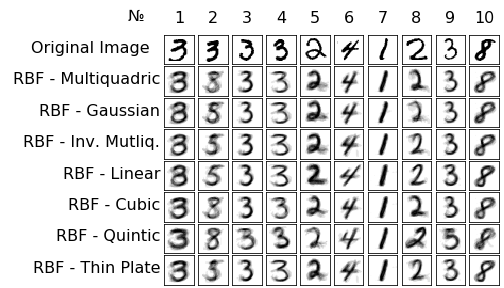}
	}
\end{figure*}

RBF interpolation performance was quite similar across all kernels, except for quintic. Refer to figure \ref{fig_cluster_rbf_all} for additional insights on RBF interpolation results. Embeddings of all shown examples except \number7 stayed in the same clusters as their nearest neighbors in the training dataset. The example \number7 was embedded in different cluster, but still it was a cluster of the same digits "2". This is an indirect indication that RBF generalized training data, rather than memorized and overfitted the examples.

\begin{figure*}
	\centering
	\caption{Cluster attribution test: power parameter for LION-tSNE and non-local IDW interpolation}
	\label{fig_power_vs_accuracy_all}
	\subfloat[Overview\label{subfig_cluster_power_vs_accuracy}]{
		\includegraphics[width=0.48\figurewidth]{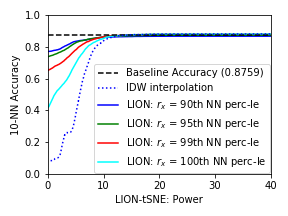}
	}
	\hfill
	\subfloat[Zoomed around baseline accuracy\label{subfig_cluster_power_vs_accuracy_zoomed}]{
		\includegraphics[width=0.48\figurewidth]{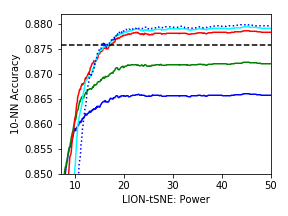}
	}	
\end{figure*}

\begin{figure*}
	\centering
	\caption{Cluster attribution test: RBF interpolation. First 10 test cases.}
	\label{fig_cluster_idw_all}
	\subfloat[Positioning. Closest training set image connected to embedding results for test image. \label{subfig_cluster_IDW_pos}]{
		\includegraphics[width=\figurewidth]{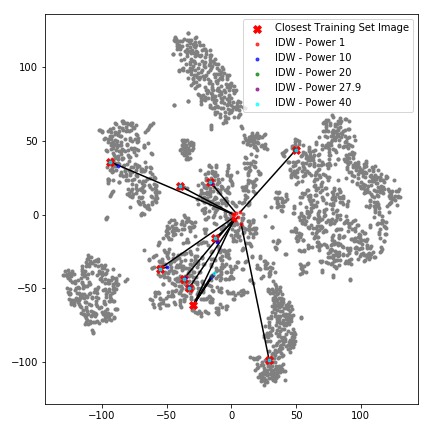}
	}
	\hfill
	\subfloat[Average 10 nearest neighbors\label{subfig_cluster_IDW_avg}]{
		\includegraphics[width=\figurewidth]{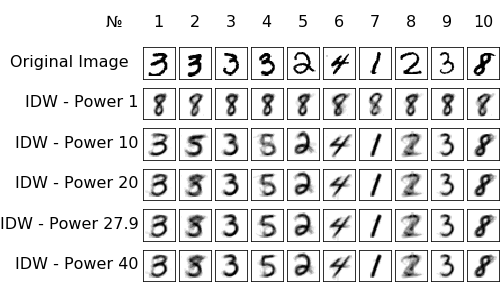}
	}	
\end{figure*}

The performance of IDW interpolation is very sensitive to correct choice of the power parameter. Refer to figure \ref{fig_cluster_idw_all} for additional insight. For power 1 most of the results are almost at the center of the plot. This effect was already described in section \ref{section_selecting_power_parameter} and depicted in \ref{fig_power_toy_example}. The power is too low, and the output is close to $mean_i(y_i)$ almost all the time. Figure \ref{fig_power_vs_accuracy_all} shows the dependency between power and 10-nearest-neighbors accuracy for cluster attribution test. Although it might look that accuracy converged at $p \approx$12, figure  \ref{subfig_cluster_power_vs_accuracy_zoomed} shows that accuracy levels off only at $p \approx 25-30$, which is the value shown by technique described in \ref{section_selecting_power_parameter}. After that increasing $p$ will produce the same accuracy with increased overfit risks. IDW shows consistent, although small, improvement even over the baseline accuracy. It should be noted that power selection metrics described in section \ref{section_selecting_power_parameter} identified good power parameters both for LION-tSNE and for non-local IDW interpolation. It also should be noted that LION-tSNE is more robust to parameter values, the reason is that if the power is too low, in local interpolation the sample is placed around the average $y$ of a neighborhood (see figure \ref{fig_power_toy_example}), and it is usually a better choice than placing it at the global average $y$ for non-local IDW interpolation.

\begin{figure*}
	\centering
	\caption{Cluster attribution test: gradient descent. First 10 test cases.}
	\label{fig_cluster_gd_all}
	\subfloat[Positioning. Closest training set image connected to embedding results for test image. For purpose of clarity only examples 2,6,7 and 10 are plotted. \label{subfig_cluster_GD_pos}]{
		\includegraphics[width=\figurewidth]{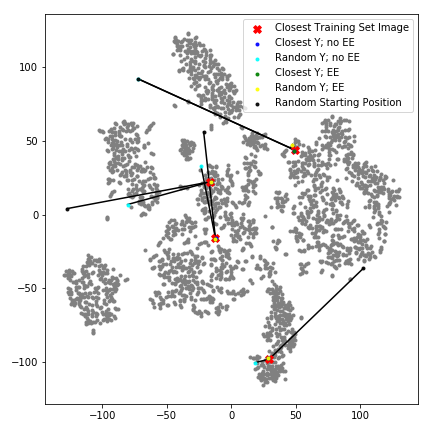}
	}
	\hfill
	\subfloat[Average 10 nearest neighbors\label{subfig_cluster_GD_avg}]{
		\includegraphics[width=\figurewidth]{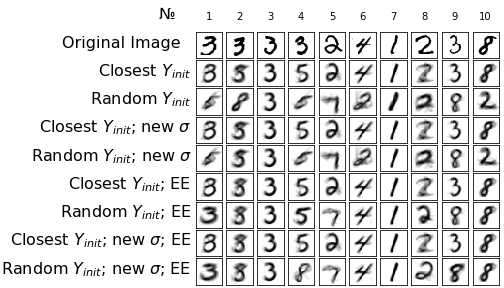}
	}		
\end{figure*}

Some of the gradient descent variations showed relatively high accuracy. The performance of gradient descent depends mainly on one choice: whether $y$ was initialized at $y_i$, corresponding to $x_i$ closest to $x$, or whether $y$ was initialized randomly. Closest initialization ensures starting close to the cost function minimum, and it takes not that many iterations to reach it. In case of random initialization gradient descent gets stuck in poor local minima. See figure \ref{subfig_cluster_GD_pos} for illustration. Early exaggeration helps to get out of poor local minima, and improves the result in case of random initialization. Recalculating $\sigma_i$ has no practical effect. Initializing at the closest $x$ seems to have particular synergy with early exaggeration.

\begin{table*} \caption{KL divergences after approximation}  \label{tab_nn_kl_div}
	\begin{tabular}{| M{0.23\textwidth} | M{0.20\textwidth}| M{0.20\textwidth}| M{0.20\textwidth} |}
		\hline
		\textbf{Configuration} & 2L; 250N; ReLu; D0.25 & 2L; 500N; ReLu; D0.5 & 1L; 500N; tanh 
		\\ \hline
		\textbf{KL divergence} & 1.3701965 & 1.4051886 & 1.1315920 
		\\ \hline
	\end{tabular}
\end{table*}

\begin{figure*}
	\centering
	\caption{Cluster attribution test: Neural Network approximation. First 10 test cases.}
	\label{fig_cluster_nn_all}
	\subfloat[Positioning. Closest training set image connected to embedding results for test image. \label{subfig_cluster_NN_pos}]{
		\includegraphics[width=\figurewidth]{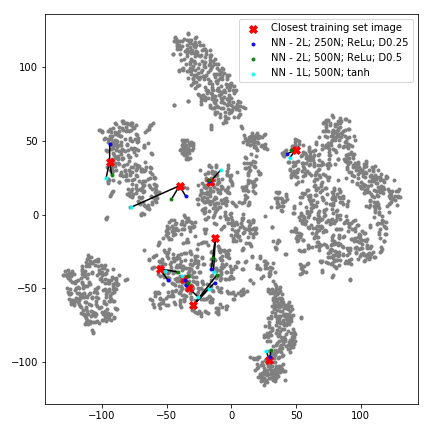}
	}
	\hfill
	\subfloat[Average 10 nearest neighbors in approximated mapping\label{subfig_cluster_NN_avg}]{
		\includegraphics[width=\figurewidth]{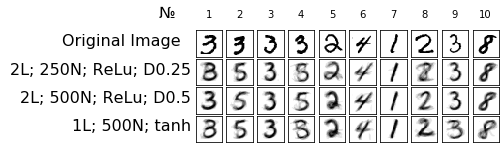}
	}
\end{figure*}

Figure \ref{fig_cluster_nn_all} shows the performance of the chosen benchmark neural network models. Neural networks approximate tSNE results, and for each model the placement of $y_i$ is slightly different (though visually barely distinguishable from original tSNE results). In \ref{fig_cluster_nn_all} for comparability all mappings $y$ are depicted in comparison to a single mapping - original tSNE mappings of $y_i$. The accuracy of neural networks with 2 is almost 10\% lower than baseline accuracy. One hidden layer model has even lower performance, which is understandable - it showed worse performance comparing to two hidden layer models on validation as well, and it was picked as a best one-hidden-layer model. It should be noted that nearest neighbors for accuracy calculations and figure \ref{subfig_cluster_NN_avg}, as well as nearest neighbors distance distribution for percentile metric in table \ref{tab_cluster_methods_comparison} were selected using corresponding approximated $y_i$, not original ones. There is also notable increase in KL divergence. However, the latter is due to approximation, not due to adding new data. KL divergences of neural network approximations (before adding any new data) are presented in table \ref{tab_nn_kl_div}.

\begin{figure*}
	\centering
	\caption{Cluster attribution test: $K$ parameter for kernelized tSNE}
	\label{fig_kernelized_k_vs_accuracy}
	\includegraphics[width=\figurewidth]{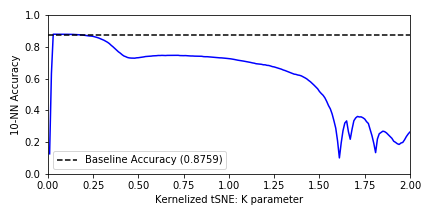}
\end{figure*}

\begin{figure*}
	\centering
	\caption{Cluster attribution test: kernelized tSNE. First 10 test cases.}
	\label{fig_cluster_kernelized_tsne_all}
	\subfloat[Positioning. Closest training set image connected to embedding results for test image. \label{subfig_cluster_kernelized_tsne_pos}]{
		\includegraphics[width=\figurewidth]{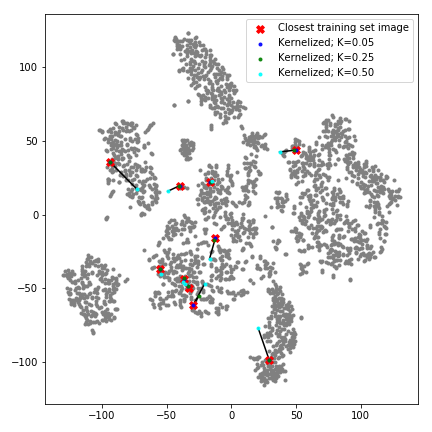}
	}
	\hfill
	\subfloat[Average 10 nearest neighbors in approximated mapping\label{subfig_cluster_kernelized_tsne_avg}]{
		\includegraphics[width=\figurewidth]{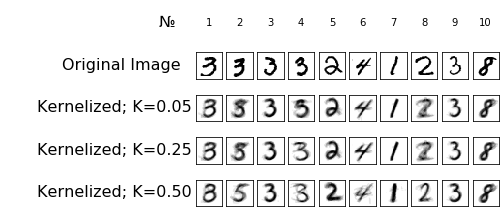}
	}
\end{figure*}

Kernelized tSNE described in \cite{gisbrecht_parametric_2015} showed near-baseline accuracy for properly chosen $K$ parameters. Performance is presented in figure \ref{fig_cluster_kernelized_tsne_all} and table \ref{tab_cluster_methods_comparison}. Figure \ref{fig_kernelized_k_vs_accuracy} shows the dependency between accuracy and parameter $K$. There is a clear window of maximum performance, after which accuracy decreases.

\begin{figure*}
	\centering
	\caption{Cluster attribution test: LION-tSNE. First 10 test cases.}
	\label{fig_cluster_lion_all}
	\subfloat[Positioning. Closest training set image connected to embedding results for test image. \label{subfig_cluster_lion_pos}]{
		\includegraphics[width=\figurewidth]{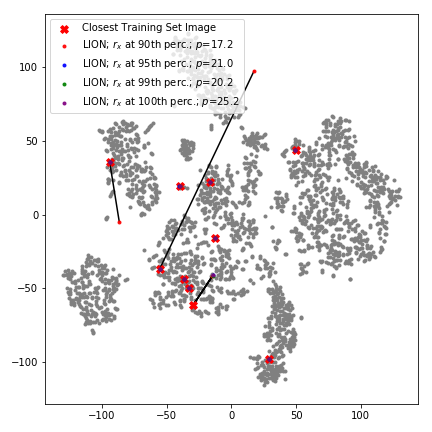}
	}
	\hfill
	\subfloat[Average 10 nearest neighbors\label{subfig_cluster_lion_avg}]{
		\includegraphics[width=\figurewidth]{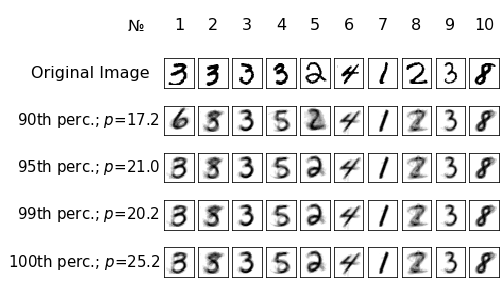}
	}		
\end{figure*}

LION-tSNE performance is presented in figure \ref{fig_cluster_lion_all} and table \ref{tab_cluster_methods_comparison}. Accuracy converged to the baseline accuracy for nearly all choices of $r_x$. Actually, figure \ref{subfig_cluster_power_vs_accuracy_zoomed} shows that LION-tSNE configurations with large $r_x$ converged to better than baseline accuracy. Figure \ref{subfig_cluster_lion_pos} illustrates the reason why small $r_x$ can cause accuracy loss: some test samples were incorrectly labeled as outliers. The probability of it happening increases with decreasing radius $r_x$. Larger radius $r_x$ converges at a higher level of $p$, but convergence results in better accuracy. This effect becomes more pronounced in higher dimensionality, so it will be explained in more details in section \ref{section_large_data}. Still it should be noted that even in worst configuration accuracy was just $\approx$ 1\% lower than baseline accuracy. LION-tSNE also showed one of the lowest KL divergences and very close distance to nearest neighbors in the cluster. Figure \ref{subfig_cluster_power_vs_accuracy_zoomed} shows some important insights on LION-tSNE performance. Also note that all chosen $r_x$ converged when power $p$ was approximately equal to the power selected by the procedure described in section \ref{section_selecting_power_parameter}, although for small $r_x$ slightly higher power seems to result in better accuracy. IDW interpolation accuracy is an upper limit here - it can be viewed as the case when $r_x \to \infty$. However, increasing radius $r_x$ over 100th percentile of nearest neighbors distance can produce accuracy improvement only by a tiny fraction. Although IDW interpolation did show better accuracy, the improvement is only around 0.0005 (0.8792 vs 0.8797 for power 50), and as we will show further, IDW interpolation is one of the least successful in dealing with outliers.

It can be safely said that in this test LION-tSNE is tied for the best performance with IDW interpolation and exceeded all other considered benchmark methods. This was one of the reasons why IDW was chosen for local interpolation in LION-tSNE - it works well when new sample $x$ is close to existing training data. However, as we shall see further, IDW interpolation is struggling to handle outliers properly. Also it should be noted that LION-tSNE is much more robust to power parameter $p$ (see figure \ref{fig_power_vs_accuracy_all}) - too low power will result in $y$ converging to the average $y$ among close neighbors in $x$, rather than to average $y$.

\subsection{Outlier Test} \label{section_outlier_test}

\begin{figure*}
	\centering
	\caption{Outlier test: first 30 test cases out of 1000}
	\label{fig_outliers_example}
	\includegraphics[width=\figurewidth]{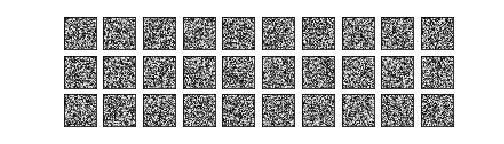}
\end{figure*}

The second test deals with outlier samples. If a new sample does not belong belongs to any local structure in $x$, it should not belong to any local structure in $y$. For MNIST dataset we generate uniform random noise, then retain 1000 examples where distance from generated $x$ to nearest neighbor $x_i$ is greater than any distance between nearest neighbors in the training set. The examples of those generated outliers are in figure \ref{fig_outliers_example}. After generating embeddings $y$ for test samples $x$, we evaluate the results using the following metrics, largely similar to the metrics of the previous test.

\textbf{Nearest neighbor distance.} This time we do expect $y$ to be an outlier. Like in previous test, we obtain distance to nearest neighbor for $x$ and determine at what percentile of the distribution $D_{NN}$ is that distance. Since $y$ should be outlier, the values of 90\% or above are acceptable. Average nearest neighbor distance is also reported - it is more meaningful comparison criteria for higher percentiles.

\textbf{Kullback-Leibler divergence} Outliers have small values of $p_{ij}$, so even if $q_{ij}$ is large (i.e. outlier was placed close to some exiting point in $y$), the increase in KL divergence will not be significant. Still the increase in KL divergence can be an additional criteria for method comparison.

Outliers test results are presented in table \ref{tab_outliers_methods_comparison}.

\begin{table*}\caption{Outliers test: methods comparison} \label{tab_outliers_methods_comparison}
	\begin{tabular}{| m{0.39\textwidth} | m{0.15\textwidth} | m{0.16\textwidth} | m{0.15\textwidth} |}
		\hline
		\textbf{Method}
		& \textbf{Distance Percentile}
		& \textbf{Distance}
		& \textbf{KL Divergence}
		\\ \hline
		\textbf{Baseline} & - & - & 1.11688
		\\ \hline
		RBF - Multiquadric & 78.785 & 4.372 & 1.12235
		\\ \hline
		RBF - Gaussian & 84.765 & 6.597 & 1.12271
		\\ \hline
		RBF - Inverse Multiquadric & 77.950 & 4.175 & 1.12245
		\\ \hline
		RBF - Linear & 83.454 & 4.844 & 1.12235
		\\ \hline
		RBF - Cubic & 74.596 & 3.882 & 1.12239
		\\ \hline
		RBF - Quintic & 99.757 & 98.044 & 1.12383
		\\ \hline
		RBF - Thin Plate & 79.766 & 4.551 & 1.12233
		\\ \hline
		IDW - Power 1 & 55.687 & 1.486 &1.12236
		\\ \hline
		IDW - Power 10 & 41.728 & 1.289 &1.12256
		\\ \hline
		IDW - Power 20 & 72.443 & 3.236 &1.12233
		\\ \hline
		IDW - Power 27.9 & 75.240 & 3.931 &1.12233
		\\ \hline
		IDW - Power 40 & 75.788 & 3.802 &1.12236
		\\ \hline
		GD - Closest $Y_{init}$ & 96.976 & 3.933 & 1.12254
		\\ \hline
		GD - Random $Y_{init}$ & 98.003 & 14.983 & 1.12264
		\\ \hline
		GD - Closest $Y_{init}$; new $\sigma$ & 96.976 & 3.933 & 1.12254
		\\ \hline
		GD - Random $Y_{init}$; new $\sigma$ & 98.003 & 14.985 & 1.12264
		\\ \hline
		GD - Closest $Y_{init}$; EE & 97.630 & 3.856 & 1.12254
		\\ \hline
		GD - Random $Y_{init}$; EE & 98.517 & 7.729 & 1.12253
		\\ \hline
		GD - Closest $Y_{init}$; new $\sigma$; EE & 97.623 & 3.852 & 1.12254
		\\ \hline
		GD - Random $Y_{init}$; new $\sigma$; EE & 98.551 & 7.724 & 1.12253
		\\ \hline
		NN - 2L; 250N; ReLu; D0.25 & 83.093 & 5.353 &1.37469
		\\ \hline
		NN - 2L; 500N; ReLu; D0.5 & 81.548 & 6.136 &1.41004
		\\ \hline
		NN - 1L; 500N; tanh & 61.333 & 2.706 &1.13739
		\\ \hline
		Kernelized; K=0.05 & 17.898 & 0.906 &1.12322
		\\ \hline
		Kernelized; K=0.25 & 84.906 & 5.140 &1.12234
		\\ \hline
		Kernelized; K=0.50 & 67.796 & 2.674 &1.12234
		\\ \hline
		\textbf{LION} -  $r_x$ at 90th perc. & \textbf{100.000} & \textbf{9.956} &\textbf{1.12385}
		\\ \hline
		\textbf{LION} -  $r_x$ at 95th perc. & \textbf{100.000} & \textbf{24.176} &\textbf{1.12387}
		\\ \hline
		\textbf{LION} -  $r_x$ at 99th perc. & \textbf{100.000} & \textbf{22.530} &\textbf{1.12385}
		\\ \hline
		\textbf{LION} -  $r_x$ at 100th perc. & \textbf{100.000} & \textbf{42.539} &\textbf{1.12383}
		\\ \hline
		
	\end{tabular}
\end{table*}

\begin{figure*}
	\centering
	\caption{Outlier test: RBF interpolation. First 10 test cases. Each mapping of a single outlier is an independent test, relative positions of outliers are not representative.}
	\label{fig_outliers_rbf}
	\includegraphics[width=0.7\figurewidth]{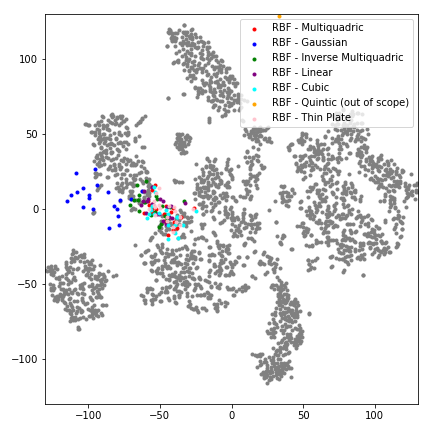}
\end{figure*}

\begin{figure*}
	\centering
	\caption{Outlier test: IDW interpolation. First 10 test cases. Each mapping of a single outlier is an independent test, relative positions of outliers are not representative.}
	\label{fig_outliers_idw}
	\includegraphics[width=0.7\figurewidth]{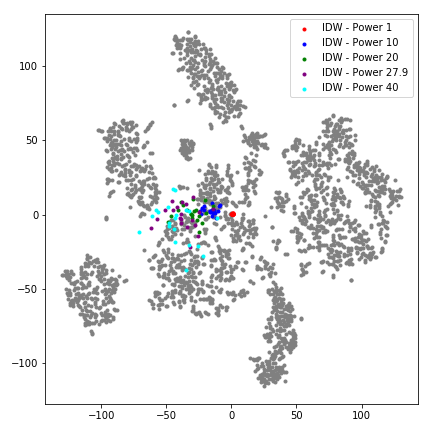}
\end{figure*}
\begin{figure*}
	\centering
	\caption{Outlier test: gradient descent. First 10 test cases. Each mapping of a single outlier is an independent test, relative positions of outliers are not representative.}
	\label{fig_outliers_gd}
	\includegraphics[width=0.7\figurewidth]{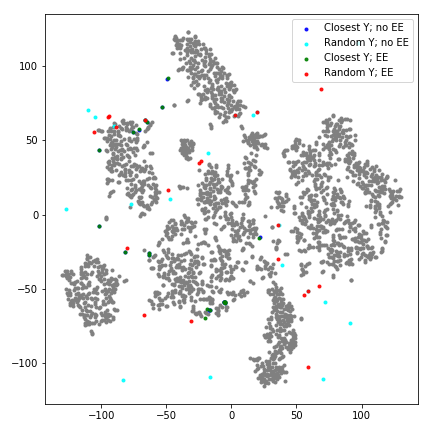}
\end{figure*}
\begin{figure*}
	\centering
	\caption{Outlier test: neural networks. First 10 test cases. Each mapping of a single outlier is an independent test, relative positions of outliers are not representative.}
	\label{fig_outliers_nn}
	\includegraphics[width=0.7\figurewidth]{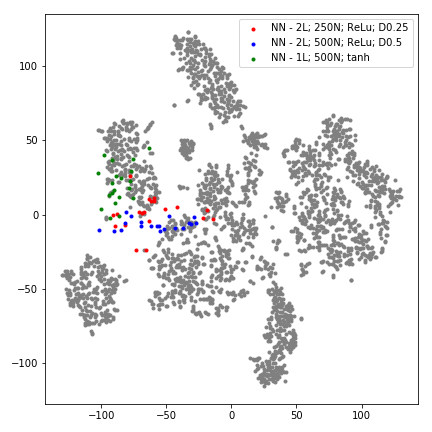}
\end{figure*}
\begin{figure*}
	\centering
	\caption{Outlier test: kernelized tSNE. First 10 test cases. Each mapping of a single outlier is an independent test, relative positions of outliers are not representative.}
	\label{fig_outliers_kernelized_tsne}
	\includegraphics[width=0.7\figurewidth]{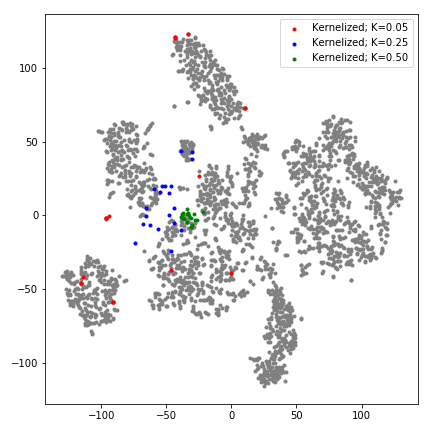}
\end{figure*}

\begin{figure*}
	\centering
	\caption{Outlier test: LION-tSNE. First 10 test cases. Each mapping of a single outlier is an independent test, overlaps are possible and relative positions of outliers are not representative. For mapping of multiple outliers at once see figure \ref{fig_outlier_placement_example}}
	\label{fig_outliers_lion}
	\includegraphics[width=0.7\figurewidth]{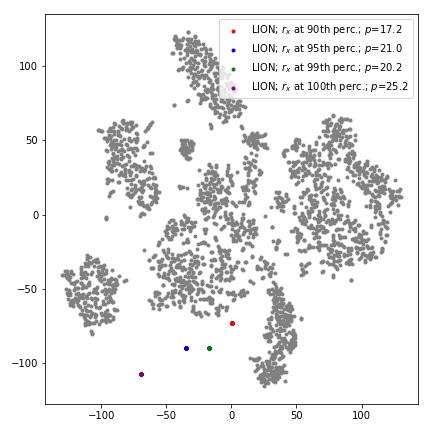}
\end{figure*}

RBF interpolation turned out to be not robust to outliers (see  figure \ref{fig_outliers_rbf}). Most outliers were placed in the existing clusters or close to them. The only exception was RBF interpolation with quintic kernel. However, this method had lower accuracy on cluster attribution test. Also it placed outliers outside of plot boundaries - it can complicate visualization.

IDW interpolation also did not show outlier robustness (see  figure \ref{fig_outliers_idw}). Interpolation with power $p=1$ resulted in all samples gathered at the center, an effect described in section \ref{section_selecting_power_parameter} and illustrated in figure \ref{fig_power_toy_example}. Larger power $p$ resulted in somewhat better performance, but still outliers overlap with existing clusters a lot, and the distances to outliers are close to the distances within the cluster.

Neural network approximation placed outliers relatively close or inside original clusters (see figure \ref{fig_outliers_nn}). It might be a subject of future work whether any other neural network configuration can achieve both high accuracy and outlier robustness.

Repeated gradient descent showed better outlier robustness than most other methods (see  figure \ref{fig_outliers_gd}), along with high accuracy in previous test. Still many outliers are indistinguishable from points in clusters. Although distance from sample to outlier is at high percentile of nearest neighbor distance, much cleaner separation could be achieved. And it should be noted that repeated gradient descent was the slowest method by far, which also limits its applicability.

\begin{figure*}
	\centering
	\caption{Outliers test: $K$ parameter for kernelized tSNE}
	\label{fig_kernelized_k_vs_distance}
	\includegraphics[width=\figurewidth]{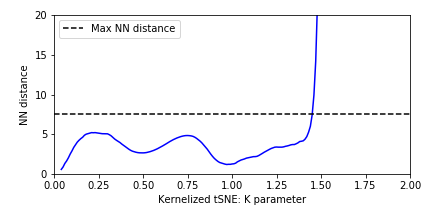}
\end{figure*}

Kernelized tSNE \cite{gisbrecht_parametric_2015} showed excellent accuracy on previous test, but in this test it placed most outliers firmly within existing clusters (see  figure \ref{fig_outliers_kernelized_tsne}). The approach is an interpolation method, and it tends to have somewhat similar struggles with outliers as all interpolation methods did. Exact influence of $K$ parameter on kernelized tSNE performance in outliers test is presented in figure \ref{fig_kernelized_k_vs_distance}.

Outlier placement test by LION-tSNE is illustrated in figure \ref{fig_outliers_lion}. LION-tSNE achieved the best outlier separation by far, and yet did not place outliers outside current plot bounds. LION-tSNE by construction places all outliers at a distance greater than the chosen threshold, so that level of performance was expected. It should be noted that in this test we generated samples that were very clearly outliers (and, it should be said, the test case definition of an outlier was consistent with LION-tSNE definition of an outlier). However, LION-tSNE performed better than any other considered method, and significantly better than the majority of considered methods. For many benchmark methods, even though outlier placement results are good on average, for each particular outlier its proper placement is not guaranteed, and figures \ref{fig_outliers_rbf}, \ref{fig_outliers_idw}, \ref{fig_outliers_gd} show examples of that - some mapped points are definitely not outliers. LION-tSNE guarantees by construction that if a sample is recognized as an outlier it will placed away from other points.

Most LION-tSNE configurations achieved both high cluster attribution accuracy and proper outlier placement - those two features are not ususally seen together on any benchmark methods, usually it was either one or the other. Moreover, LION-tSNE in both cases showed the best performance among compared methods. In cluster attribution test LION-tSNE was tied for the best accuracy several other methods. For outlier placement LION-tSNE showed the best results by far. Some methods came close to LION-tSNE in outlier placement test in terms of percentile (several gradient descent variations, RBF interpolation with quintic kernel, and gradient descent), however, actual nearest neighbor distance show that LION-tSNE achieves much better separation between clusters and outliers.

One more prominent method, to which LION-tSNE was not yet compared, is parametric tSNE \cite{maaten_learning_2009}. This method approximates tSNE using restricted Boltzmann machines (RBMs). RBMs are the earliest version of deep learning models \cite{hinton_fast_2006}, and they often require a lot of data to be trained. Despite extensive grid search we could not find configuration that provided good performance after training 2500 samples in 30 dimensions. So, in order to provide fair comparison, we used dataset to which it was originally applied: 784-dimensional MNIST dataset with 60000 samples. The next section describes application of LION-tSNE to larger and higher-dimensional datasets using parametric tSNE \cite{maaten_learning_2009} as a comparison benchmark.

\subsection{Increasing Data Set Size and Dimensionality}\label{section_large_data}

This section aims to evaluate LION-tSNE performance when the size and the dimensionality of the dataset is increased. For evaluation we used 60000 samples from MNIST dataset in 28x28 = 784 dimensional space. Test set consists of 10000 samples, and those samples were used for cluster attribution test. As a comparison benchmark we used parametric tSNE with 3 hidden layers having 500,500, and 2000 nodes respectively. This layer configuration was used for evaluation by parametric tSNE author \cite{maaten_learning_2009}, source code of parametric tSNE and the training/test dataset was made available by the original algorithm author. For outlier robustness test we used the same generated outliers like in previous section. In order to give both algorithms equal footing, we use $Y$ values found by parametric tSNE with mentioned configuration and the perplexity of 30 as the ground truth for LION-tSNE interpolation.

\begin{figure*}
	\centering
	\caption{Cluster attribution test on high-dimensional dataset: power parameter for LION-tSNE and parametric tSNE}
	\label{fig_power_vs_accuracy_large}
	\includegraphics[width=\figurewidth]{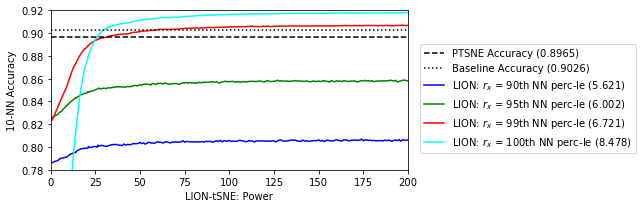}
\end{figure*}

The accuracy of cluster attribution test and its dependence on power parameter $p$ are presented in figure \ref{fig_power_vs_accuracy_large}. Baseline accuracy is now defined as the average 10 nearest neighbors accuracy in $y$ space for all 60000 samples of the training set. Note that the effect described in section \ref{section_cluster_attribution_test} became more pronounced - smaller radius $r_x$ converges faster, but to lower accuracy. Still for large $r_x$ values the accuracy was consistently better than the baseline. The converged performance is mentioned in the table \ref{tab_large_dataset_p}.

So, why different $r_x$ converge to different accuracy? The table \ref{tab_large_dataset_p} provides results and intermediate calculations. Consider what happens with an embedding $y$ of a test sample $x$ for a very large $p$. In IDW interpolation for $p \to \infty$ the embedding $y \to y_i$, where $y_i$ is the embedding of closest nearest neighbor $x_i$ (see figure \ref{fig_power_toy_example} for illustration). In LION-tSNE it depends on the radius $r_x$ whether that nearest neighbor is used for local approximation or not. However, if two nearest neighbors are within $r_x$, the radius $r_x$ no longer plays a role, the result is determined: $y \to y_i$. Consider the accuracy for $r_x \to \infty$ and $p \to \infty$: it can be calculated straightforwardly by finding nearest neighbor $x_i$ for every test case $x$, then picking $y_i$ and calculating its 10 nearest neighbor accuracy (including the sample $y_i$ itself in the neighbor list, but using expected label of $y$ as the class label). That value is 0.9200, and that's what accuracy values should converge to when $p \to \infty$ and if at least two nearest neighbors are in $r_x$ for all test samples. The latter is not always the case, that's why most $r_x$ converged to even higher accuracy values. The accuracy is averaged not for all test samples, but only for those that had at least 2 nearest neighbors within smaller $r_x$. The samples with larger number of close neighbors are in denser clusters and tend to have higher than average 10 nearest neighbor accuracy.

However, as further lines of table \ref{tab_large_dataset_p} illustrate, for small $r_x$ higher accuracy for non-outliers is outweighed by larger number of outliers. Accuracy can be viewed as a weighted sum of 3 components: $acc = f_{2N}*acc_{2N} + f_{1N}*acc_{1N} + f_{O}*acc_{O}$. The value $f_{2N}$, $f_{1N}$ and $f_{O}$ represent fraction of test cases that have 2 or more neighbors in $r_x$, that have one a neighbor in $r_x$ (and still were not considered an outlier) and the outliers (with one or zero neighbors) respectively. The values $acc_{2N}$, $acc_{1N}$ and $acc_{O}$ represent corresponding accuracy values respectively. Note that only one component of accuracy depends on the power $p$ - it is $acc_{2N}$. As table \ref{tab_large_dataset_p} shows, the component $f_{1N}*acc_{1N}$ is not significant due to very low number of samples with 1 outlier nearest neighbor in the training set. The component $f_{O}*acc_{O}$ also have insignificant impact due to low 10 nearest neighbor accuracy for outlier samples. The component $f_{2N}*acc_{2N}$ has dominating impact. For lower $r_x$ higher accuracy for non-outliers is outweighed by larger number of outliers, and as a result final accuracy values are smaller. That is why lower $r_x$ converge to worse accuracy. This also shows importance of proper selection of radius $r_x$ for LION-tSNE.

\begin{table*} \caption{LION-tSNE accuracy estimation}  \label{tab_large_dataset_p}
	\begin{tabular}{| M{0.42\textwidth} | M{0.09\textwidth}| M{0.09\textwidth}| M{0.09\textwidth}| M{0.09\textwidth}|}
		\hline
		\textbf{$NN_x$ percentile} &90 & 95 & 99 & 100
		\\ \hline
		\textbf{Radius $r_x$} &5.62 & 6.00 & 6.72 & 8.48
		\\ \hline
		\makecell{\textbf{Power (by cross-validation)}}& 13.8 & 16.5 & 21.1 & 26.6
		\\ \hline
		\makecell{\textbf{Accuracy} \\ \textbf{(cross-validated $p$)}}& 0.7954 & 0.8457 & 0.8889 & 0.8978
		\\ \hline
		\makecell{\textbf{Accuracy} \\ \textbf{($p$ = 200)}}&0.8063&0.8583&0.9069&0.9181		
		\\ \hline
		\textbf{Accuracy $acc_{2N}$ ($p \to \infty$, $\ge$ 2 neighb. in $r_x$)}& 0.9436 & 0.9365 & 0.9258 & 0.9200
		\\ \hline
		\textbf{Fraction $f_{2N}$}& 0.8390 & 0.9108 & 0.9805 & 1.0000
		\\ \hline
		\textbf{$acc_{2N} * f_{2N}$}& 0.79168 & 0.85293 & 0.9078 & 0.9200
		\\ \hline		
		\textbf{Accuracy $acc_{O}$ (outlier, <2 neighb. in $r_x$)}& 0.0522 & 0.0420 & 0.0293 & -
		\\ \hline
		\textbf{Fraction $f_{O}$}& 0.1522 & 0.0848 & 0.0191 & 0
		\\ \hline
		\textbf{$acc_{O} * f_{O}$}& 0.0079 & 0.0036 & 0.0006 & 0		
		\\ \hline
		\textbf{Accuracy $acc_{1N}$ (1 neighb. in $r_x$, not outlier)}& 0.8011 & 0.0.6500 & 0.2000 & - 
		\\ \hline
		\textbf{Fraction $f_{1N}$}& 0.0088 & 0.0044 & 0.0004 & 0
		\\ \hline
		\textbf{$acc_{O} * f_{O}$}& 0.0079 & 0.0036 & 0.0006 & 0				 							
		\\ \hline
		\makecell{\textbf{Accuracy} \\ \textbf{($p \to \infty$)}}&0.8067&0.8594&0.9084&0.9200	
		\\ \hline
	\end{tabular}
\end{table*}

\begin{figure*}
	\centering
	\caption{Power parameter selection on high-dimensional dataset}
	\label{fig_power_validation_large}
	\includegraphics[width=0.7\figurewidth]{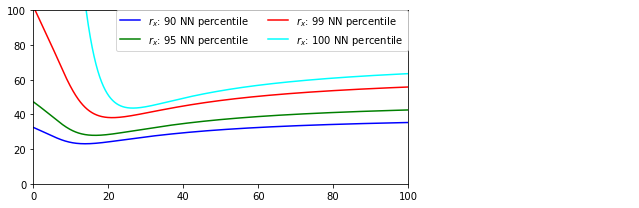}
\end{figure*}

In order to illustrate that power selection procedure described in section \ref{section_selecting_power_parameter} still works for higher dimensionality and larger dataset, consider figure \ref{fig_power_validation_large}. Power values that result in minimum of the validation function are presented in table \ref{tab_large_dataset_p}, accuracy corresponding to the chosen power is also presented in table \ref{tab_large_dataset_p}. It should still be noted that it is a heuristic, not an exact solution, and better performance might be achievable (here selected power values resulted in $\approx$1-2\% accuracy loss). Note that although for this test case high power $p$ results in higher accuracy, setting too high power is not a good solution - embedding $y$ of any new sample $x$ will be barely distinguishable from embedding $y_i$ of its nearest training set neighbor $x_i$ (see figure \ref{fig_power_toy_example}).

For outlier test we used similar outlier samples like in the test described in section \ref{section_outlier_test}.

\begin{table*} \caption{LION-tSNE performance on larger datasets with higher dimensions}  \label{tab_large_dataset_summary}
	\begin{tabular}{| M{0.37\textwidth} | M{0.08\textwidth}| M{0.08\textwidth}| M{0.08\textwidth}| M{0.08\textwidth}| M{0.09\textwidth}|}
		\hline
		\multirow{2}{*}{\textbf{Parameter}}&
		\multicolumn{4}{|c|}{\textbf{LION-tSNE: $r_x$ percentile}} &
		\multirow{2}{*}{\textbf{PtSNE}}
		\\ \cline{2-5}
		{} & \textbf{90} & \textbf{95} & \textbf{99} & \textbf{100} & {}
		\\ \hline
		\textbf{Accuracy (cross-val. $p$)}& 0.7954 & 0.8457 & 0.8889 & 0.8978 & \multirow{3}{*}{0.8965}
		\\ \cline{1-5}
	    \textbf{Accuracy ($p$=50)} & 0.8040 & 0.8547 & 0.9013 & 0.9112 & {}
		\\ \cline{1-5}
		\textbf{Accuracy ($p \to \infty$)} &0.8067&0.8594&0.9084 & 0.9200	& {}
		\\ \hline
		\textbf{Outlier NN distance} & \multicolumn{4}{|c|}{23.36} & 0.29
		\\ \hline
		\makecell{\textbf{Outlier NN distance} \\ \textbf{percentile}}& 100 & 100 & 100 & 100 & 64
		\\ \hline									    	
	\end{tabular}
\end{table*}

The resuts of both cluster atribution test and outlier robustness test is presented in table \ref{tab_large_dataset_summary}. The accuracy of LION-tSNE and parametric tSNE is approximately equal, and exact results might depend on configuration. However, like for smaller dataset, LION-tSNE achieved very good outlier separation. Both desired properties of LION-tSNE - high accuracy and clean outlier separation - scaled well with increased dataset size and dimensionality. 

The next section provides complexity analysis for LION-tSNE and benchmark algorithms and concludes evaluation.

\subsection{Complexity Analysis}

\begin{table*}\caption{Complexity: methods comparison} \label{tab_complexity_comparison}
	\begin{tabular}{| m{0.35\textwidth} | M{0.2\textwidth} | M{0.22\textwidth} |}
		\hline
		\textbf{Method}
		& \textbf{Upfront Costs}
		& \textbf{Runtime Complexity}
		\\ \hline
		RBF Interpolation & O($N^3K^3$) & O($NK$)
		\\ \hline
		IDW Interpolation & None & O($NK$)
		\\ \hline
		Gradient Descent & None & O($N^2K$)
		\\ \hline
		Neural Networks & Varies & Not $f(N)$
		\\ \hline
		Kernelized tSNE & O($N^3K^3$) & O($NK$)
		\\ \hline
		Parametric tSNE & Varies & Not $f(N)$
		\\ \hline
		\textbf{LION} & O($NK$) & O($NK$)
		\\ \hline
		
	\end{tabular}
\end{table*}

LION-tSNE was tied for highest accuracy and shown the best outlier separation. However, another comparison factor is the complexity of the algorithm. High complexity can make very accurate algorithm unusable, and low complexity can make less accurate algorithm the preferred choice. Complexity analysis is summarized in table \ref{tab_complexity_comparison} and described in more details below. In this analysis we take into account not only the number of training samples, but the number of original dimensions $K$. The dimensionality of reduced space $d$ is 2 in almost all practical cases, it will be treated as constant. We assume that $N \gg K$, so complexity like O($N+K$) is equivalent to O($N$), and O($N^2 + NK$) is equivalent to O($N^2$). However, complexity like O($NK$) is not equivalent to O($N$). 

Building RBF interpolation function requires solving system of linear equations (SLE) to obtain $\lambda_i$ (formula \ref{eq_RBF_fit}). Therefore, it takes O($N^3$) of upfront calculations to construct interpolation function. In order to build the matrix for SLE it requires O($N^2$) distance computations in $K$ dimensions, but its complexity O($N^2K$) is subsumed by O($N^3$). Once interpolation function is built, it takes O($NK$) to use it for any new $x$. The same is applicable for kernelized tSNE.

IDW interpolation does not require precomputations. Building interpolation function comes at no upfront cost. It takes O($NK$) to obtain value for new input $x$: determine all $N$ inverse distances in $K$ spaces, normalize weights and calculate weighted sum. 

Repeated gradient descent also does not require any upfront calculations. However, the practical complexity to obtain $y$ for new $x$ can vary a lot depending on how many gradient descent steps are required. Straightforward implementation of tSNE algorithm is has quadratic complexity in number of points \cite{maaten_visualizing_2008} (if dimensionality is taken into account, $O(N^2K)$), and it is the upper bound of repeated gradient descent complexity. Practically at runtime gradient descent worked much slower than any other method.

Neural network approximation and restricted Boltzmann machines for parametric tSNE \cite{maaten_learning_2009} require upfront training using backpropagation. Once they are trained, it takes several matrix multiplications and activation functions to determine the final value. So, both upfront and runtime complexity highly depend on the model structure: number of layers, number of nodes in each layer, etc. The main advantage of those methods is that runtime complexity does not depend on $N$ - it can be important for very large training sets. Considering that input layer has $K$ nodes and output layer has $d$ nodes, complexity depends on $K$ and $d$, but reporting it as merely O($K$) can be very misleading, there are too many other factors involved.

The complexity of LION-tSNE is close to the complexity of IDW interpolation, upon which it is based. It takes O($NK$) to determine the proper neighborhood for local interpolation, and building the interpolation itself takes O($N_{neighb}K$), which has an upper bound of O($NK$). Unlike original IDW, there is also specific upfront cost related to obtaining the coordinates for outlier positioning. Outliers are placed in special free cells (see section \ref{section_outlier_placement}). The number of cells to build grows exponentially with the dimensionality of $y$ space, but the dimensionality of $y$ space is usually 2 and here it is treated as constant. If a straightforward approach is used, for each cell it takes $N$ distance comparisons in $K$ dimensions O($NK$) to determine what training data (if any) are in the cell. Once potential coordinates are determined, they can be retrieved randomly from the pool at a constant cost. Another computation (which can be calculated upfront) is determining the coordinates of additional cells to place outliers. Again the complexity is exponential of number of reduced dimensions $d$. If separation distance in $r_y$ or close proximity radius $r_{close}$ are determined as a percentile of nearest neighbor distance in $y$ space, it takes O($N^2d^2$) to build that distance distribution, but it can be skipped if necessary. To summarize, unavoidable upfront costs are O($NK$)

In summary, LION-tSNE is tied for the lowest runtime complexity with most interpolation methods. Linear complexity shows decent scaling to growing $N$ and $K$. LION-tSNE does require some upfront calculations related to outlier placement. If necessary these costs can be avoided by simplifying outlier placement procedures - placing outliers outside main plot area only, and skip detecting free cells inside the main plot area. This solution will drop upfront complexity to $const(N, K)$ (assuming maximum and minimums of $y$ for each dimension were calculated before), but might come at a cost of growing plot area.

\section{Discussion} \label{section_discussion}

	In evaluation LION-tSNE achieved both good accuracy and outlier robustness. One could argue that LION-tSNE worked well on the cases where it was designed to work well. On another hand "a sample belongs to a certain cluster" and "a sample is an outlier" are usually the cases of most practical interest. In addition, borderline cases are difficult to evaluate, so comparing the performance and judging what method is a better fit is complicated.
	
	In order to understand applicability of LION-tSNE, consider several alternative methods and their use cases. Consider a dynamic data scenario, where at any time a sample can appear, disappear or change. The scenario can be viewed also as a series of data snapshots, but in case of asynchronous updates the changes between each snapshot will be in one sample only. There are several possible approaches to the challenge:
	
	\textbf{- Re-run tSNE.} This is a most straightforward approach, and most likely we will see the same data structures (along with new clusters, if those appeared). However, the changes for each particular data sample will not be visible. If there are many data snapshots over time, or if there is a stream of new samples, then re-run of tSNE is not feasible approach - each snapshot visualization will look completely different, and each new or modified sample will result in a completely new visualization (can be especially important when data change asynchronously - one sample at a time). Moreover, running tSNE again is often much slower than using interpolation or approximation. To summarize, tSNE is highly useful for static snapshots, but it requires extension to handle dynamic data scenarios. Re-running tSNE is feasible if the only question is which data structures are appearing and disappearing, and if we can afford to analyze and compare the snapshots manually.
	
	\textbf{- Use LION-tSNE} (or any other interpolation approach over tSNE data). These approaches can handle scenarios where there is a stream of new samples over time (including both new samples and modified samples), or when there is a stream of data snapshots. It will be immediately visible for each sample, if that sample stayed in place, or moved insignificantly, or moved to a different cluster, or (in case of LION-tSNE) if the sample became an outlier. Also it will be visible if some cluster is disappearing, and it might be visible if some new cluster is being formed. Building visualization animations is possible. Handling asynchronously changing data does not present any additional challenges. To summarize, those approaches can handle a lot of dynamic data scenarios.
	
	\textbf{- Approximation of tSNE results.} The approach is close to the previous one and mainly shares the same pros and cons. There is a potential for increased KL divergence and outlier robustness might be a challenge, but on the upside runtime complexity does not depend on the number of training samples $N$. It can be highly beneficial for large $N$. The ways to improve LION-tSNE and reduce the complexity in term of $N$ are discussed below in this section.

	\textbf{- Using tSNE time series visualization approaches} (like points on a single tSNE plot \cite{nguyen_m-tsne:_2016} or like a series of plots \cite{rauber_visualizing_2016}). Those approaches are designed to visualize dynamic data as a whole. Handling the stream of completely new samples (unrelated to previous ones) is out of scope. Visualizing the facts on the level of exact samples like "sample $x_i$ stays in place/moves/changes cluster at time $t$" is also out of scope. To summarize, those methods were built for a different use case, and the preference should depend on exact task.
	
	There are several special cases that might need explicit attention and improvements. Most of them are especially important for the data that changes over time. For illustration consider a moving data point $x(t)$ with its embedding $y(t)$.
	
	First special case is when $x(t)$ leaves the $r_x$-neighborhood of all raining samples and becomes an outlier. In that case there will be a sudden change in position. In our opinion, this is acceptable behavior: it will attract attention to the point, but it won't mislead the user while exploring the data.
	
	Another special case is when $x(t)$ leaves or enters $r$-neighborhood of certain training point $x_i$. It is another source of possible discontinuity: situation is equivalent to forcing interpolation weight $w_i(x)$ to zero (or from zero to some value) instantly. If it is not acceptable, it can be alleviated as follows: each participating inverse distance $\|x-x_i\| ^{-p}$ can be further multiplied by any continuous function that gradually transitions from 1 when $\|x-x_i\| = 0$ to 0 when $\|x-x_i\| \ge r$.  For example, it can be a triangular or trapezoid-shaped function. Only then, after multiplication, the weights are normalized. As a result, when transitioning out of $r$-neighborhood of some training point $x_i$, the weight $w_i(x)$ reaches exactly zero when $\|x-x_i\| = r$, and continuity is maintained.
	
	Third special case is moving between similar clusters. For example, in figure \ref{fig_mnist_tsne_original} consider two clusters that correspond to digit "1" (large lower center cluster and smaller cluster on upper left). Consider a data sample $x(t)$, which gradually transitions from some point in one cluster $x_i$ to a point in another cluster $x_j$. And consider a situation when $x(t)$ is exactly between clusters: it did not become an outlier while transitioning, and there are members of both clusters in its $r$-neighborhood in original space. Then local interpolation will place this point somewhere between those clusters, i.e. in completely different cluster that has nothing in common with current point $x(t)$. It can be highly misleading during data exploration, and this is another case when discontinuity is better behavior. Plausible solution is to limit the distance between neighbors in $y$ space - if the point $x_i$ is in $r$-neighborhood of point $x(t)$, but its distance between $y_i$ and other local neighborhood points exceed some threshold, then this point is a candidate for exclusion from local interpolation. We can remove candidates starting from most distant point $argmax(\|x-x_i\|)$, until there are no candidates for exclusion. If this method is applied, then $y(t)$ will make a sudden transition from one cluster to another, when the number of neighbors in the second cluster reaches some critical point.
	
	There is still a room for improvement of outlier handling. At the moment outlier location is chosen randomly, but it can be chosen to better reflect if a point $x$ is close to some training sample (e.g. by choosing the outlier position which is closest  to that sample). Also what happens if outlier changes its position over time? If the change of position is insignificant, i.e. $|x_{new} - x_{old}| \le r_x$, and the point did not stop being an outlier, then it can be reflected by randomly moving outlier at close proximity to its old position. It will both indicate that position has changed, but the change is not significant. Another question is: what if new cluster is being formed? Though grouping outliers together allows detecting that, exact performance of LION-tSNE in that scenario is yet to be determined.
	
	Next section summarizes future work directions and concludes the article.

\section{Conclusion and Future Work} \label{section_conclusion}

Some possible improvements and future work opportunities were already mentioned in section \ref{section_discussion}.  However, there is several major future work direction that were not mentioned before, and those can have major impact on LION-tSNE performance.

Perhaps, the most impactful future work direction is using fast nearest neighbors search. Straightforward approach takes O($N$) comparisons to determine neighbors in a certain radius around new sample $x$. It can be a problem when $N$ is large. Search for nearest neighbors in a certain fixed radius is a recognized problem on its own \cite{bentley_survey_1975}. Also fixed radius nearest neighbors search can be approximated with fast K nearest neighbors search techniques \cite{cunningham_k-nearest_2007} with sufficient number of neighbors and additional check whether the distance is small enough. Another option is to use orthogonal range search \cite{berg_computational_2008}, i.e. search for neighbors in a box, rather than in a sphere.

And, of course, application of LION-tSNE to large variety of practical tasks can help to identify further room for improvement of the algorithm.

In summary, tSNE algorithm is highly useful for exploring and visualizing high dimensional datasets. In this article we addressed the challenge of using tSNE to visualize dynamic data streams. We proposed, analyzed, implemented and evaluated LION-tSNE - a novel approach based on local interpolation and special outlier handling. The approach was compared with large set of benchmark approaches, and it was tied for highest accuracy and showed the best outlier separation by far. There is still room for future research, but LION-tSNE can already be applied to a variety of practical data analytics tasks.

\section*{Acknowledgments}
This research was supported by BGL BNP Paribas and Alphonse Weicker Foundation. We'd like to thank Anne Goujon and Fabio Nozza from BGL BNP Paribas for their assistance.

\bibliographystyle{plain}
\bibliography{references}

\end{document}